\documentclass[sigconf]{acmart}

\setcopyright{none}
\renewcommand\footnotetextcopyrightpermission[1]{} % removes footnote with conference information in first column

\acmConference[ESEC/FSE 2023]{The 31st ACM Joint European Software Engineering Conference and Symposium on the Foundations of Software Engineering}{11 - 17 November, 2023}{San Francisco, USA}

\usepackage[ruled,linesnumbered]{algorithm2e}
\usepackage{algorithmic}
\usepackage{enumitem}
\usepackage{dsfont}
\usepackage{color}

\AtBeginDocument{%
  \providecommand\BibTeX{{%
    \normalfont B\kern-0.5em{\scshape i\kern-0.25em b}\kern-0.8em\TeX}}}

\acmConference[ESEC/FSE 2023]{The 31st ACM Joint European Software Engineering Conference and Symposium on the Foundations of Software Engineering}{11 - 17 November, 2023}{San Francisco, USA}

\newcommand{{\methodname}}{TIN}

\begin{document}

\title{Automated Testing and Improvement of Named Entity Recognition Systems}



\author{Boxi Yu}
\email{boxiyu@link.cuhk.edu.cn}
% \orcid{0000-0001-5213-7189}
\affiliation{
  \institution{School of Data Science, The Chinese University of Hong Kong, Shenzhen (CUHK-Shenzhen)}
  % \city{Shenzhen}
  % \state{Guangdong}
  \country{China}
}

\author{Yiyan Hu}
\email{yiyanhu@link.cuhk.edu.cn}
% \orcid{0000-0001-5213-7189}
\affiliation{
  \institution{School of Science and Engineering, The Chinese University of Hong Kong, Shenzhen (CUHK-Shenzhen)}
  % \city{Shenzhen}
  % \state{Guangdong}
  \country{China}
}

\author{Qiuyang Mang}
\email{qiuyangmang@link.cuhk.edu.cn}
% \orcid{0000-0001-5213-7189}
\affiliation{
  \institution{School of Data Science, The Chinese University of Hong Kong, Shenzhen (CUHK-Shenzhen)}
  % \city{Shenzhen}
  % \state{Guangdong}
  \country{China}
}

\author{Wenhan Hu}
\email{wenhanhu@link.cuhk.edu.cn}
% \orcid{0000-0001-5213-7189}
\affiliation{
  \institution{School of Data Science, The Chinese University of Hong Kong, Shenzhen (CUHK-Shenzhen)}
  % \city{Shenzhen}
  % \state{Guangdong}
  \country{China}
}

\author{Pinjia He}
\authornote{Pinjia He is the corresponding author.}

\email{hepinjia@cuhk.edu.cn}
% \orcid{0000-0001-5213-7189}
\affiliation{
  \institution{School of Data Science, The Chinese University of Hong Kong, Shenzhen (CUHK-Shenzhen)}
  % \city{Shenzhen}
  % \state{Guangdong}
  \country{China}
}

\begin{abstract}
We develop the first black-box automated testing tool for testing general named entity recognition systems, i.e., named entity and nested named entity.

\end{abstract}

%    <concept>
%        <concept_id>10002978.10003022.10003023</concept_id>
%        <concept_desc>Security and privacy~Software security engineering</concept_desc>
%        <concept_significance>500</concept_significance>
%        </concept>
%  </ccs2012>

% \ccsdesc[500]{Security and privacy~Software security engineering}




\begin{abstract}

Named entity recognition (NER) systems have seen rapid progress in recent years due to the development of deep neural networks.
These systems are widely used in various natural language processing applications, such as information extraction, question answering, and sentiment analysis.
However, the complexity and intractability of deep neural networks can make NER systems unreliable in certain circumstances, resulting in incorrect predictions.
For example, NER systems may misidentify female names as chemicals or fail to recognize the names of minority groups, leading to user dissatisfaction.
To tackle this problem, we introduce \textit{{\methodname}}, a novel, widely applicable approach for automatically testing and repairing various NER systems.
The key idea for automated testing is that the NER predictions of the same named entities under similar contexts should be identical.
The core idea for automated repairing is that similar named entities should have the same NER prediction under the same context.
We use {\methodname} to test two SOTA NER models and two commercial NER APIs, i.e., Azure NER and AWS NER.
We manually verify 784 of the suspicious issues reported by {\methodname} and find that 702 are erroneous issues, leading to high precision (85.0\%-93.4\%) across four categories of NER errors: omission, over-labeling, incorrect category, and range error.
For automated repairing, {\methodname} achieves a high error reduction rate (26.8\%-50.6\%) over the four systems under test, which successfully repairs 1,056 out of the 1,877 reported NER errors.

\end{abstract}

\maketitle

\section{Introduction}
% intro and significance of ner
Named entity recognition (NER) systems have become popular in recent years and are widely used to enhance natural language processing (NLP) applications such as information retrieval, machine translation, and text classification.
With the development of the Internet, mountains of data accumulate every day.
Common Crawl,\footnote{https://commoncrawl.org/} a nonprofit organization that crawls the web and freely provides its archives and datasets to the public, has collected 380 TB of data and 3.15 billion pages by October 2022.
As of 30 December 2022, there are 6,594,544 articles in the English Wikipedia and it contains over 4 billion words~\cite{wiki_size}.
Being able to identify the semantics of interest in unstructured texts, NER systems play a significant role in several downstream applications.
For instance, chatbots like ChatGPT and Google's Bard employ NER to identify and categorize entities in user queries, enhancing response accuracy, while in finance, NER algorithms sift through reports and online mentions, extracting and classifying data for in-depth profitability analysis and real-time stock market trend monitoring.\footnote{https://www.techtarget.com/whatis/definition/named-entity-recognition-NER}
News providers, such as publishing houses, can harness NER to sift through and categorize their abundant daily content.
By detecting crucial entities like people, organizations, and places in articles, they can seamlessly tag and organize them, optimizing content discovery for readers.\footnote{https://towardsdatascience.com/named-entity-recognition-applications-and-use-cases-acdbf57d595e}
% For example, news publishers can use NER to suggest similar articles based on the entities discussed within them.
% For example, a NER system can be used to identify the named entities in a document, which would be further used to extract information about those entities for a knowledge base or to answer questions about the entities.
% add citation of 003 challenge

% performance of ner systems
Though achieving shining F1-Score on multiple NER benchmarks (e.g., CoNLL03~\cite{sang2003introduction} and OntoNotesv5~\cite{weischedel2013ontonotes}), current NER systems are far from perfect, and errors produced by these systems largely dissatisfy the users and could lead to detrimental influence.
In NER systems, errors pertain to either the incorrect delineation of named entities within a text corpus or the misclassification of such identified entities into inappropriate categories.
In Table~\ref{tab:ner_implication}, we present multiple instances of NER errors, encompassing four categories: omission, over-labeling, incorrect category, and range error.
For example, in the first sentence, the name \textit{"Paul"} is clearly identifiable as a person.
However, the NER system fails to recognize any word as a named entity, leading to an omission error.
The inaccuracies stemming from NER errors can reverberate across a broad spectrum of domains.
For example, in the report by Mishra et al.~\cite{mishra2020assessing}, some NER models are better at identifying White names across all datasets with higher confidence compared with other demographics, such as Black names.
In addition, Zhao et al.~\cite{zhao2022comprehensive} found that some NER systems are prone to identifying female names as chemicals, and most NER systems perform better on male-related data than female-related data.
% These NER errors could damage various downstream tasks (e.g., knowledge base generation) because NER systems serve as the basis for them and NER errors might propagate.

\begin{table*}[h]
\caption{Example of the NER errors}
\centering
\resizebox{.99\linewidth}{!}{
\begin{tabular}{@{}clcc@{}}
\toprule
\textbf{Error Type} &
  \multicolumn{1}{c}{\textbf{Sentence}} &
  \textbf{Predicted Entities} &
  \textbf{Target Entities} \\ \midrule
\textit{Omission}&
  \begin{tabular}[c]{@{}l@{}}Sir \underline{Paul}'s command of the stage is so casual that he makes it look easy (\textbf{Flair-Ontonotes}). \end{tabular} &
  \begin{tabular}[c]{@{}c@{}} {\textcolor{red}{NULL}} \\  \end{tabular} &
  \begin{tabular}[c]{@{}c@{}} \textcolor{blue}{{[}"Paul", \texttt{PER}{]}} \end{tabular} \\ \midrule
\textit{Over-labeling}&
  \begin{tabular}[c]{@{}l@{}} \underline{Elon Musk} is having a similar effect on the platform (\textbf{Azure}). \end{tabular} &
  \begin{tabular}[c]{@{}c@{}}{[}"Elon Musk", \texttt{PERSON}{]} \\ \textcolor{red}{{[}"Platform", \texttt{LOCATION}{]}} \\  \end{tabular} &
  \begin{tabular}[c]{@{}c@{}}{[}"Elon Musk", \texttt{PERSON}{]} \end{tabular} \\ \midrule
\textit{Incorrect Category}&
  \begin{tabular}[c]{@{}l@{}} \underline{Norrie} believes securing \underline{Unesco} status could offer new opportunities \\ in sustainable tourism and branding of local produce, while at the \\ same time highlighting the environmental value of the peatland (\textbf{Flair-Conll}). \end{tabular} &
  \begin{tabular}[c]{@{}c@{}} {[}"Norrie", \texttt{PER}{]} \\ \textcolor{red}{{[}"Unesco", \texttt{MISC}{]}}\\  \end{tabular} &
  \begin{tabular}[c]{@{}c@{}}{[}"Norrie", \texttt{PER}{]} \\ \textcolor{blue}{{[}"Unesco", \texttt{ORG}{]}} \end{tabular} \\ \midrule
\textit{Range Error}& 
  \begin{tabular}[c]{@{}l@{}}\underline{Det Supt Rance} said the investigation remained active (\textbf{AWS}). \end{tabular} &
  \begin{tabular}[c]{@{}c@{}} \textcolor{red}{{[}"Det", \texttt{PERSON}{]}} \\ \textcolor{red}{{[}"Supt Rance", \texttt{PERSON}{]}} \\  \end{tabular} &
  \begin{tabular}[c]{@{}c@{}} \textcolor{blue}{{[}"Det Supt Rance", \texttt{PERSON}{]}} \end{tabular} \\ \bottomrule
\end{tabular}
\label{tab:ner_implication}
}
\end{table*}

\begin{table*}
    \centering
    \caption{NER systems with different standards}
    \vspace{-2ex}
    \resizebox{1.0\linewidth}{!}{
    \begin{tabular}{cc}
    \hline
    NER Systems & \multicolumn{1}{c}{ NER Categoreis } \\
    \hline
    Flair-CoNLL & \textbf{PERSON}, \textbf{ORGANIZATION}, \textbf{LOCATION}, MISCELLANEOUS NAMES  \\
    % \hline
    Flair-Ontonotes & \textbf{PERSON}, \textbf{ORGANIZATION}, \textbf{LOCATION}, CARDINAL, DATE, EVENT, FAC, GPE, LANGUAGE, LAW, MONEY, NORP, ORDINAL, PERCENT, PRODUCT, QUANTITY, TIME, WORK-OF-ART   \\
    % \hline
    Azure NER & \textbf{PERSON}, \textbf{ORGANIZATION}, \textbf{LOCATION}, PERSONTYPE, EVENT, PRODUCT, SKILL, ADDRESS, PHONENUMBER, EMAIL, URL, IP, DATETIME, QUANTITY \\
    % \hline
    AWS NER & \textbf{PERSON}, \textbf{ORGANIZATION}, \textbf{LOCATION}, COMMERCIAL ITEM, DATE, EVENT, OTHER, QUANTITY, TITLE \\
    \hline
    % MS Azure API & 88.13   \\
    % \hline
    \end{tabular}
    \label{tab:ner_category}
    }
    \vspace{-2ex}
\end{table*}

Although the reliability of NER systems is of great importance, there is a dearth of general methods for automatically testing and repairing various NER systems since it is quite challenging.
First, unlike traditional code-based software, NER systems are mainly based on deep neural networks with millions of parameters.
Therefore, many testing techniques that perform well on code-based software may not be suitable for testing NER systems.
% Secondly, the current methods for \bx{just work here!}
Second, it is laborious to manually construct the test oracle, i.e., large amounts of text data with labels indicating the presence of named entities, such as people and organizations.
% if we want to evaluate NER systems on real-world data.
Third, NER systems have multiple standards for identifying and categorizing named entities in text, which further compounds the difficulty of designing a general method for testing and repairing various NER systems.

Traditional approaches for repairing AI-based systems either adopt data augmentation~\cite{yu2022automated, he2020structure, gupta2020machine} or algorithm optimization.
Thus, these approaches need manual labeling of the data or modification of the model architecture, which often incurs a high cost.
In addition, these traditional methods need to access the model in a white-box manner, while a black-box testing and repair approach would be much more general.
% which is difficult to be adapted to repair the closed-sourced systems.

To address the challenges, we propose \textit{{\methodname}}, the first approach for automatically testing and repairing NER systems, which is applicable to NER systems with various standards.
% The core idea of automated testing is that the NER predictions under similar contexts should be the same. 
% Based on this idea,
We design three transformation schemes for test case generation, including similar sentence generation, structural transformation, and random entity shuffle.
Meanwhile, we design the corresponding metamorphic relations for each of the transformation schemes.
% \hyy{We may separate this sentence into two}.
% For automated repairing, the core idea is that similar named entities should have the same NER prediction under the same context.
Then we construct the test inputs consisting of the original sentence, the mutant sentence, and their NER predictions.
The test input is reported as a suspicious issue if it does not satisfy the corresponding metamorphic relation.
After receiving the suspicious issues from the automated testing, our location algorithm would be used to locate the suspicious entity that is prone to NER errors.
For each suspicious entity, we use BERT~\cite{devlin2018bert} to generate similar named entities and a novel repairing algorithm to predict the correct NER category of the suspicious entity by leveraging the prediction of similar entities under the same context.
% of the generated named entities.
% It is worth noticing that {\methodname} is applicable to NER systems with different standards, which addresses the third challenge.

We apply {\methodname} to test two SOTA NER models from Flair~\cite{flairAPI} (a widely-used NLP library), and two commercial APIs from Azure and AWS.
The two SOTA models Flair-CoNLL and Flair-Ontonotes are trained with the NER standard of CoNLL03~\cite{sang2003introduction} and OntoNotesV5 \cite{weischedel2013ontonotes}, respectively.
For the commercial APIs, we denote Azure Cognitive Services for Language as "Azure NER", and AWS Sagemaker as "AWS NER" for simplicity.
To verify the effectiveness of {\methodname}, we manually inspect part of the results for testing and repairing.
{\methodname} successfully reports 702 erroneous issues out of 784 suspicious issues with a high precision ranging from 85.0\% to 93.4\% on the four NER systems under test.
The detected NER errors in the erroneous issues include omission, mislabeling, incorrect category, and range error.
In terms of NER repairing, {\methodname} improves predictions with a high rate ranging from 48.1\%. to 52.2\%, while only downgrading predictions in only a low ratio from 12.1\% to 19.5\%.
{\methodname} also exhibits a high error reduction rate (26.8\%-50.6\%) on the four NER systems, demonstrating its effectiveness in repairing NER systems and reducing NER errors. All the source code and data in this work will be released for reuse.

This paper makes the following main contributions:
\begin{itemize}
    \item It introduces {\methodname}, a novel, widely-applicable approach for automatically testing and repairing NER systems.
    \item {\methodname} provides three transformation schemes for test case generation and the corresponding metamorphic relations for NER error detection. 
    \item {\methodname} contains a novel repairing algorithm that can effectively fix the reported NER errors.
    \item It presents the evaluation of {\methodname} against SOTA models and commercial APIs, achieving a high precision of 85.0\%-93.4\% for reporting erroneous issues, and a high error reduction rate of 26.8\%-50.6\% for repairing the NER errors.
    % \item It discusses the diverse NER categories reported by {\methodname} and how {\methodname} sucessfully repairs these NER errors.
\end{itemize}

\section{Preliminaries}

% \begin{table*}
%     \centering
%     \caption{NER systems with different standards}
%     \resizebox{1.0\linewidth}{!}{
%     \begin{tabular}{cc}
%     \hline
%     NER Systems & \multicolumn{1}{c}{ NER Categoreis } \\
%     \hline
%     Flair-CoNLL & \textbf{PERSON}, \textbf{ORGANIZATION}, \textbf{LOCATION}, MISC  \\
%     % \hline
%     Flair-Ontonotes & CARDINAL, DATE, EVENT, FAC, GPE, LANGUAGE, LAW, \textbf{LOCATION}, MONEY, NORP, ORDINAL, \textbf{ORGANIZATION}, PERCENT, \textbf{PERSON}, PRODUCT, QUANTITY, TIME, WORK-OF-ART   \\
%     % \hline
%     Azure NER & \textbf{PERSON}, PERSONTYPE, \textbf{LOCATION}, \textbf{ORGANIZATION}, EVENT, PRODUCT, SKILL, ADDRESS, PHONENUMBER, EMAIL, URL, IP, DATETIME, QUANTITY \\
%     % \hline
%     AWS NER & COMMERCIAL ITEM ,DATE, EVENT, \textbf{LOCATION} \textbf{ORGANIZATION}, OTHER, \textbf{PERSON}, QUANTITY, TITLE \\
%     \hline
%     % MS Azure API & 88.13   \\
%     % \hline
%     \end{tabular}
%     \label{tab:ner_category}
%     }
% \end{table*}

\subsection{Named Entity Recognition}
% Natural language processing (NLP) has achieved great process in recent decades that focuses on enabling computers to understand, interpret, and generate human language.
NER systems are essential tools in NLP because they allow machines to extract structured information from unstructured text.
NER systems are used in a variety of applications, including information extraction, machine translation, and question answering.
For example, a NER system might be used to extract the names of people, organizations, and locations from a news article.
This information can then be used to build a database of entities and their relationships, which can be queried and analyzed to gain insights or to perform tasks such as information retrieval or summarization.
% They are also helpful for tasks such as entity disambiguation, where a system must determine which entity is being referred to in a given context.
Overall, NER systems are significant because they enable machines to understand and make sense of large volumes of unstructured text, which is an increasingly important task in today's digital world.
% Traditionally, approaches to repairing AI-based systesm use data augmentation~\cite{yu2022automated, he2020structure, gupta2020machine}or adversarial training to fix the bugs or 

NER task has multiple standards, e.g., the NER dataset CoNLL03 \cite{sang2003introduction} and Ontonotesv5~\cite{weischedel2013ontonotes} have different NER standards.
In addition, the four NER systems have different NER standards, which only share the common categories of \texttt{PERSON} (\texttt{PER}), \texttt{LOCATION} (\texttt{LOC}), and \texttt{ORGANIZATION} (\texttt{ORG}) (in Table~\ref{tab:ner_category}).
The various identification standards further exacerbate the difficulty of manual test oracle construction.
% the  the design of the test oracle.

\subsection{Constituency Parser}
A constituency parser is a natural language processing tool that is used to analyze and understand the syntactic structure of sentences.
It works by breaking down a sentence into its constituent parts, known as constituents, and arranging them in a tree-like structure known as a phrase structure tree or a constituency tree.
The tree represents the syntactic relationships between the words in a sentence and the grammatical roles they play.
% Constituency parsers use context-free grammars (CFGs)~\cite{cfg} to analyze sentences and produce phrase structure trees.
In the process of analyzing sentences and producing phrase structure trees, constituency parsers use context-free grammars (CFGs~\cite{cfg}).
CFGs are a set of rules that specify how words can be combined to form phrases and sentences.
The parser uses these rules to identify the constituents of a sentence and the relationships between them.

In {\methodname}, we adopt the constituency parser implemented in Stanford CoreNLP~\cite{stanford_corenlp} to parse the sentences.
In the constituency parser, a non-terminal node is a node in the phrase structure tree that represents a syntactic category, such as a noun phrase (NP), verb phrase (VP), and sentence (S).
Non-terminal nodes are used to represent the internal structure of the sentence and the relationships between its constituents.
A terminal node is a leaf node in the phrase structure tree that represents an individual word or punctuation mark in the sentence.
Terminal nodes are used to represent the basic building blocks of the sentence, and they do not have any children.

% \vspace{-2ex}
\subsection{Problem Definition}

\subsubsection{NER Testing}
We denote the NER systems under test as $N$, and the input sentence as $s$.
{\methodname} adopts metamorphic testing to alleviate the test oracle problem.
$N(s)$ represents the output of the NER systems, which contains a list of NER predictions with the corresponding NER categories, e.g., \texttt{PERSON} or \texttt{ORGANIZATION}.
{\methodname} transforms the original sentence $s$ to obtain the mutant sentence $s'$ through transformation $s'=T(s)$ by utilizing the structure of the sentence $s$ and the NER output $N(s)$.
We then design the corresponding metamorphic relation (MR) for the NER output pair $(N(s), N(s'))$, and report the $(N(s), N(s'))$ along with $(s, s')$ as a suspicious issue if the metamorphic relation is unsatisfied.
% \hyy{Should we create a notation for the metamorphic relation?}.

\subsubsection{NER Repairing}
NER repairing aims to repair the NER errors and gives the correct NER prediction.
We design a repairing system $R$ to repair the NER predictions of the suspicious issues reported by the testing part of {\methodname}.
We repair both the NER prediction of the original sentence and the mutant sentence and obtain the repaired NER predictions $R(s)$ and $R(s')$.

\section{Approach and Implementation}
% \bx{do not forget to explain the choice of hyper-parameters}

\begin{figure*}[th]
 \centering
 \includegraphics[width=0.82\linewidth]{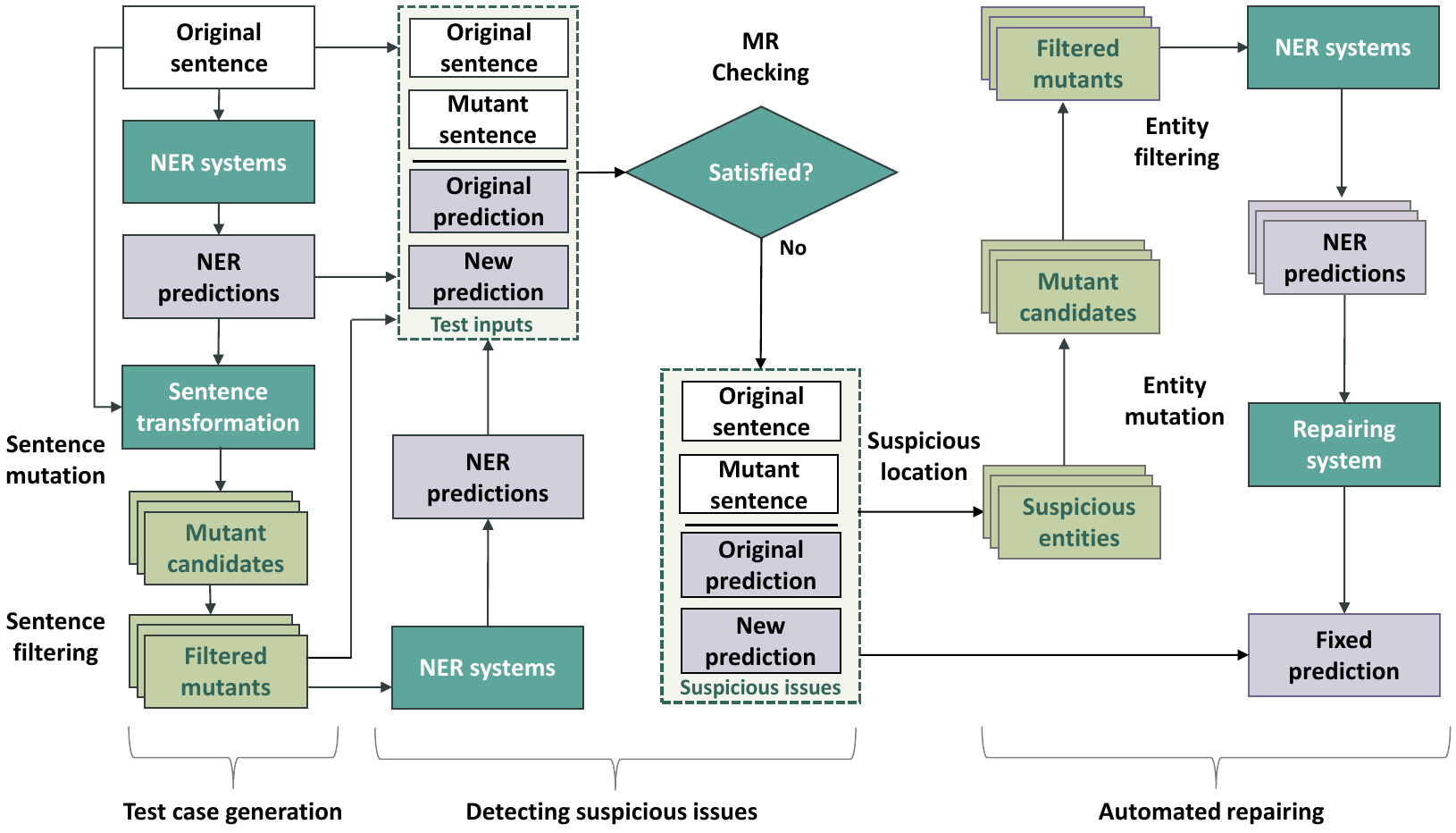}
 \caption{Overview of {\methodname}}
 \label{fig:framework}
\end{figure*}

\subsection{Overview}
The overview framework of {\methodname} is shown in Fig~\ref{fig:framework}, which consists of three main components, which are test case generation, detecting suspicious issues and automated repairing, respectively.

% *** \bx{introductino of the overall framework}
The first part of {\methodname} is test case generation.
{\methodname} first feeds the original sentences into the NER systems and gets the NER prediction of the original sentences, which are used to assist the automated generation of test cases.
After generating multiple mutant candidates, we use sentence filters to filter out low-quality sentences.
% , where the details of the filters will be explained in Section~.
The filtered mutants are then used for automated testing.

The second part of {\methodname} is detecting suspicious issues.
We construct the test inputs, which consist of the sentence and the NER predictions of the original and mutant sentences.
We design the metamorphic relation for each of the transformation schemes and report the suspicious issues that may contain NER errors if the metamorphic relations are violated.

The third part of {\methodname} is automated repairing.
After we obtain the suspicious issues from the second part of {\methodname}, we start to locate the suspicious entities that may lead to the NER errors.
% We mask the suspicious entities in the sentence needed to be repaired, then 
We use BERT~\cite{devlin2018bert} to generate mutant named entities with similar semantics to replace the suspicious entities.
We then filter them through entity filtering to obtain the filtered mutants.
We feed the filtered mutants to the NER systems and obtain a list of NER predictions for repairing the suspicious entities.
In practice, we develop a scoring system to leverage the entity predictions to repair the suspicious entities.

% The words with similar semantics should have the same named entity category in a NER system, e.g., when we replace the named entity "America" with "China" in the sentence "America is a great country" to generate the sentence "China is a great country", the NER type for the location of the word "America" in the sentence should also be labeled as "location".

% \textbf{1) A}

% \textbf{2) A}

% \subsection{Preparation for generating sentences}

\subsection{Automatically Generating Test Cases}

% \begin{figure*}
%  \centering
%  \includegraphics[width=0.77\linewidth]{Images/architecture.pdf}
%  \caption{Overview of {\methodname}}
%  \label{fig:framework}
% \end{figure*}
To automatically generate test cases, we design three transformation schemes to generate mutant sentences from the original sentence.
The transformation schemes consist of: (1) similar sentence generation, (2) structural transformation, and (3) random entity shuffle.
We will introduce each of these transformation schemes and their implementations in the following. 
% \hyy{I think we should mention that we aim to generate sentences with similar context with the original sentence}

% \begin{algorithm}\small
%     \caption{Algorithm for automatic testing on sentences \label{alg:reg}} 
%     \KwData{
%     ss: a sentence input;}
%  \KwResult{suspicious\_issuessuspicious\_issues}
%     suspicious\_issues = \mathrm{List}()suspicious\_issues = \mathrm{List}()\\
%     p_{ori} = \mathrm{model}(x)p_{ori} = \mathrm{model}(x)\\
%     new\_sentences = \mathrm{Generator}(s, p_{ori})new\_sentences = \mathrm{Generator}(s, p_{ori})\\
%     new\_sentences = \mathrm{Naturalness\_filter}(new\_sentences)new\_sentences = \mathrm{Naturalness\_filter}(new\_sentences)
%     \For{each sentence and its metadata s_t, m_ts_t, m_t \in\in new\_sentencesnew\_sentences}
%     { 
%         flag = \mathrm{False}flag = \mathrm{False}\\
%         p_{new} = \mathrm{model}(s_t)p_{new} = \mathrm{model}(s_t)\\
%         flag = \mathrm{Check\_MR}(p_{ori}, p_{new}, m\_t)flag = \mathrm{Check\_MR}(p_{ori}, p_{new}, m\_t)\\
%         \If{flagflag}
%             {\mathrm{Add}(susipicious\_issues, (s, p_{ori}, s_t, p_{new}))\mathrm{Add}(susipicious\_issues, (s, p_{ori}, s_t, p_{new}))\\}
%     }
%     \Return suspicious\_issuessuspicious\_issues\\
% \end{algorithm}

\subsubsection{Similar Sentence Generation}
The core idea of generating similar sentences is to substitute the words or phrases in the sentences with the ones that have similar semantics.
In this procedure, we avoid modifying named entities of the original sentences predicted by the NER systems.
We implement two methods for similar sentence generation, including token-level and phrase-level similar sentence generation, which are explained as below.

\textbf{1) Token-level similar sentence generation:}
After we obtain the NER prediction of the original sentence $N(s)$,
% \hyy{We first tokenized the original sentence and select candidate tokens to be perturbed according to the pos-tag of the tokens (the candidate words can be verbs or adjectives that do not affect entity prediction), then}
we mask the token that does not belong to a named entity in turn (one word being masked each time) and feed the masked sentence into BERT to generate a list of candidate words.
% \hyy{BERT MLM} 
% We use BERT to generate cont<ext-similar words to replace the original word in the original sentence, 
% \textcolor{blue}{We choose the top-K words of the candidate words with the highest predictive logits and use the top-K words to substitute the original word to generate mutant sentences.}
We select the top-K most predictive candidate words and use them to replace the original ones and generate mutant sentences.
During this process, we only use verbs and adjectives as candidate tokens to avoid grammatically strange or incorrect sentences.
In addition, we ensure the token to be replaced keeps the identical part-of-speech, by using the NLTK's POS tagging tool~\cite{nltkAPI}.
% \hyy{You can move the NLTK part to the front, as I want to change the paragraph structure a bit}

\begin{figure}
 \centering
 \includegraphics[width=0.8\linewidth]{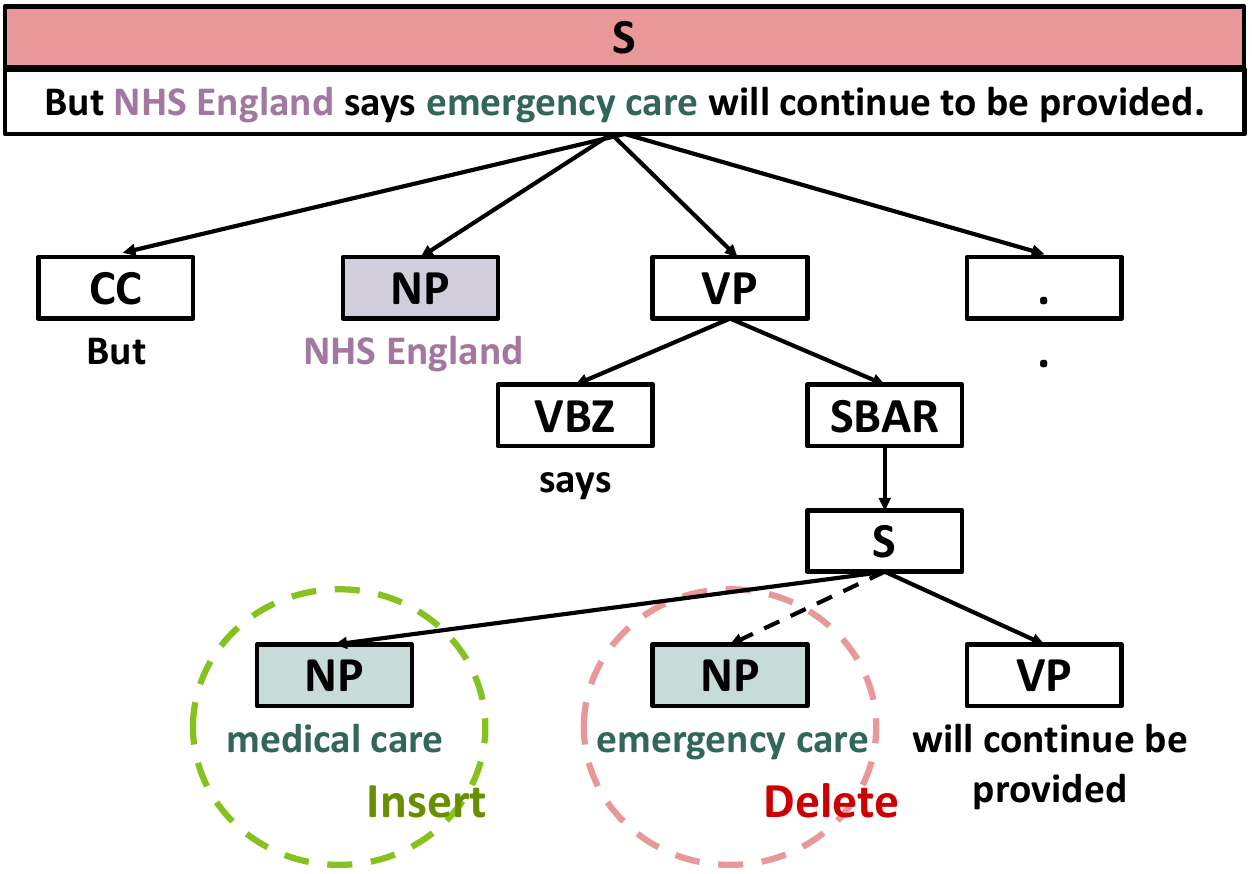}
 \caption{Phrase-level similar sentence generation}
 \label{fig:synonyms}
\end{figure}

\textbf{2) Phrase-level similar sentence generation:}
Besides token-level replacement, we design a phrase-level sentence transformation scheme by utilizing the constituency parser.
Specifically, we only modify the noun phrase to avoid grammar errors.
As shown in Fig.~\ref{fig:synonyms}, we use a constituency parser to parse the original sentence and find NP nodes where there are no other NP nodes in their subtree, i.e., the NP nodes with the smallest unit.
Then we use sense2vec~\cite{trask2015sense2vec} to generate similar noun phrases to replace the content in the NP nodes we find in turn (one NP node being replaced each time).
As shown in Fig.~\ref{fig:synonyms}, the NP node with "\textit{emergency care}" has been replaced by the NP node with "\textit{medical care}". 
Finally, we obtain a list of mutant sentences, each with one noun phrase replaced.

\subsubsection{Structural Transformation}

The general idea of structural transformation is changing the syntactic structure of the original sentence as well as ensuring the new sentences have a similar context to the original sentence.
Specifically, we provide three transformation implementations of sentence syntactic structure to transform the declarative sentence into the interrogative sentence.

We first use a constituency parser to parse the original sentence and obtain a parse tree.
 Then we find the sentence node S that satisfies the CFGs grammar of "$\text{S}\rightarrow \text{NP}\ \text{VP}$", and check the first child VP node of S that modifies the NP node.
 % There are several situations of the first child VB node of the VP node, where we use different transformation methods for each of them in the following.
 There are different ways to handle the first child VB node of the VP node, and we use a specific transformation method for each case.
 % \bx{need to ask pj about the preliminary intro of the constituency parsing.}
% Then we change the structure of the SS node to transform the declarative sentence into an interrogative sentence.
% If the declarative sentence satisfies the pattern of <subject><verb><(object)>, there are two situations:

\begin{figure}
 \centering
 \includegraphics[width=1\linewidth]{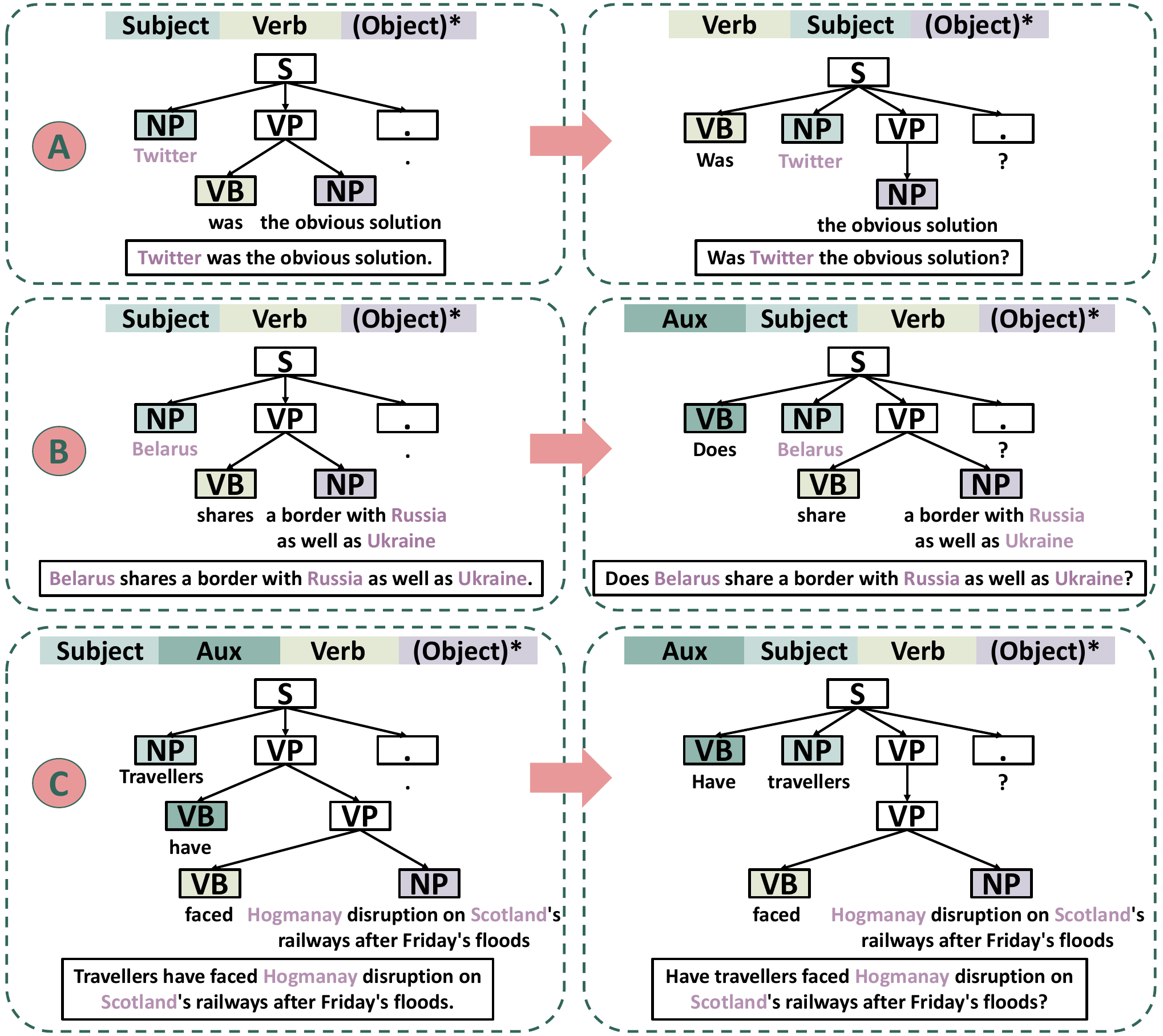}
 \caption{Structural transformation}
 \label{fig:structure}
\end{figure}

\textbf{1)} If the VB node is a "be verb" that serves as a "normal verb" and directly modifies a noun phrase, e.g., the sentence "\textit{He is a student.}", we will transform the sentence structure from <subject><verb><(obj\\ect)*> to <verb><subject><(object)*>, and get a sentence "\textit{Is he a student?}".
Please notify we use * here to represent that the object is not necessary.
% In this case, an NP node is a right neighbor node of a VB node, as shown in Fig.~\ref{fig:structure} (A).
As shown in Fig.~\ref{fig:structure} (A), we will move the VB node to the position of the first child of node S.
As a result, the declarative sentence "\textit{Twitter was the obvious solution}" is transformed into the interrogative form "\textit{Was twitter the obvious solution}".

\textbf{2)} If the VB node is a "normal verb" that is not a "be verb", e.g., the sentence with the transitive verb "\textit{I eat a burger}" or the sentence with the intransitive verb "\textit{He cried}", we will transform the sentence structure from <subject><verb><(object)*> to <Aux><subject><ver\\b><(object)*>.
As shown in Fig.~\ref{fig:structure} (B), we will insert a VB node with an "auxiliary verb" at the position of the first child of node S.

\textbf{3)} If the VB node is an "auxiliary verb", e.g., the sentence "\textit{He has faced floods}", we will transform the sentence structure from <subject><aux><verb><(object)*> to <aux><subject><verb><(obj\\ect)*>.
As shown in Fig.~\ref{fig:structure} (C), we will move the VB node to the position of the first child of node S.

After the movement or insertion, we replace the "." at the end of the sentence with "?".

\subsubsection{Random Entity Shuffle}
We propose to use random entity shuffle to generate mutant sentences and test whether the NER predictions keep the same.
% \hyy{I think we need to mention that shuffling entities of the same type has little affect on the sentence context }
% \hyy{for the word list of the original sentence: }
We first use the NER systems to obtain the NER predictions and replace the original named entities with the placeholders of their own type, e.g., <\texttt{ORG}> and <\texttt{PER}>.
Then we randomly shuffle the order of the named entities of the same category within the sentence and place the named entities into the corresponding placeholders to generate mutant sentences.

For example, the sentence "\textit{Spotify, Apple Music, and Deezer all said the track was their top performer of the year, beating competition from Ed Sheeran, Drake, and Taylor Swift.}" will be modified to "\textit{<\texttt{ORG}>, <\texttt{ORG}>, and <\texttt{ORG}> all said the track was their top performer of the year, beating competition from <\texttt{PER}>, <\texttt{PER}>, and <\texttt{PER}>}".
After we random shuffled and filled named entities for each NER category respectively, mutant sentences such as 
"\textit{Apple Music, Spotify, and Deezer all said the track was their top performer of the year, beating competition from Ed Sheeran, Taylor Swift, and Drake.}" will be generated.

\subsection{Filters to Improve the Quality of Test Cases}\label{sec:filters}
To improve the quality of the test cases generated by {\methodname}, we adopt a semantic filter to ensure the semantic similarity between the original sentence and mutant sentence, and a syntactic filter to filter out the sentences that are prone to grammar errors.
\subsubsection{Semantic Filter}
In automated testing, we have two transformation methods for similar sentence generation, including token-level and phrase-level substitution.
As the same named entity may have different NER categories under various contexts, we need to ensure that these two transformation methods only change the semantics of the sentence subtly.
% In automated repairing, we also use BERT to generate the mutant entities to replace the original named entity.
% To ensure the mutant entities have similar semantics with the original named entity.
In practice, we use BERT to obtain the context-aware embedding of the word or phrase and compute the cosine similarity between the mutant and the original one.
To avoid ambiguity, we denote $\mathrm{BERT\_MLM}$ as the function we use to generate words by using the BERT masked language model and $\mathrm{BERT\_Embedding}$ as the function that we use BERT to compute the context-aware embedding.
The context-aware embedding $h$ of the word $w$ in a sentence $s$ is denoted as $h=\mathrm{BERT\_Embedding}(w,s)$.
For a phrase that contains multiple words, we use the average of the context-aware embedding vectors in the phrase as its context-aware embedding.
Given a context-aware embedding vector $h$ and another context-aware embedding vector $h'$, the semantic similarity is defined as:
\begin{equation}
    \label{equ:cosine similarity}
    \mathrm{CosSim}(\boldsymbol{h}, \boldsymbol{h'}) = \frac{\boldsymbol{h^T} \cdot \boldsymbol{h'}}{\Vert\boldsymbol{h}\Vert\Vert\boldsymbol{h'}\Vert},
\end{equation}
which is the cosine similarity between $h$ and $h'$.

% \textcolor{blue}{Specifically, as replacing a token or phrase with its antonyms also has a subtle change for NER predictions of the sentence, e.g, changing "good" to "bad", we use its absolute value rather than the cosine distance in the Equ~\ref{equ:cosine similarity}.}

The semantic filter will filter out the mutant sentences where the semantic similarity $\mathrm{CosSim}$ is less than a threshold $SThreshold$.
Specifically, we set $SThreshold$ as 0.65 for token-level replacement and phrase-level replacement based on experience, which achieves good performance on four NER systems under test.
% \hyy{Should I mention how many sentences are filtered out in this step?}
% Specifically, we use different thresholds for testing and repairing of {\methodname} due to the difference of the target to substitute, i.e., word or phrase that are not named entity for testing, and named entity for repairing.
% In the experiment, we set $SThreshold$ as 0.65 for similar sentence generation and 0.45 for generating similarly named entities. \bx {the value needs checking}

\subsubsection{Syntactic Filter}
When we transform the original sentence to generate mutant sentences, there may include grammar errors, punctuation errors, or rarely used expressions.
% \hyy{The new sentence should be in the distribution of "normal sentence" in the real world}
% Therefore, it is significant to filter out these unnatural test cases, which rarely appear in the real world and are treated as noises~\cite{}.
After the transformation of the sentences, it is significant to filter out these test cases, which rarely appear in the real world.
To this end, we use the syntactic evaluator SynEval implemented by AEON to compute the syntactic score of our mutant sentences, which evaluates the naturalness through the perplexity of the Pre-trained Language model.
Specifically, we calculate the difference in the syntactic score between the original sentence $s$ and the mutant sentence $s'$, which is defined as:
\begin{equation}
    \epsilon = \mathrm{SynEval}(s) - \mathrm{SynEval}(s').
\end{equation}
We adopt the syntactic filter for all the transformation approaches, and use a threshold $SynThreshold$ to filter out the low-quality mutant sentences with $\epsilon$ greater than $SynThreshold$.
We use grid search to find the best $\epsilon$ value between 0 and 0.05, balancing quantity and quality of test cases.
For Flair-CoNLL, which only contains four NER categories, we set $\epsilon$ as 0.02 for structural transformation, and 0.01 for other sentence transformation.
For the other three NER systems, which have much more NER categories, we set $\epsilon$ as 0 for all transformation methods to achieve stricter filtering.
% \bx{We set different threshold for different transformation methods and NER systems, more details will be shown in the supplement.} \hyy{we use the $\epsilon$ since we want "the new sentence is at least almost as natural as the original sentence".}
% We set $SynThresho\\ld$ as 0.02 for token-level similar sentence generation and structural transformation, and set $SynThreshold$ as 0.01 for phrase-level similar sentence generation and random entity shuffle. \bx{need adjusting the hyper-parameter}

% of our mutant sentences and filter out the sentences with scores lower than a threshold.

% Filter 1: Word embedding similarity
% Filter 2: AEON

\subsection{Detecting Suspicious issues}
After we generate the mutant sentences and use the filters to select the test cases with high quality, we begin to construct the test inputs for {\methodname}.
The test inputs contain the original and mutant sentences and their predictions by the NER systems.
As shown in Fig.~\ref{fig:framework}, we examine the metamorphic relations of the test inputs.
We check if the first metamorphic relation, $MR_1$, is met for generating similar sentences.
$MR_1$ requires: 
% \hyy{Maybe we can discuss this part in the following way: 1. expect the original prediction and new prediction are the same 2. However, the new word substitution may cause new entity 3. we design two kinds of MR 4. list two MRs}

\begin{equation}
    N(s,e)=N(s',e), \forall e,\ s.t., e\in s\ \mathrm{and}\ e \in s',
\end{equation}
where $N(s,e)$ and $N(s',e)$ represents the NER prediction of $e$ in the original sentence $s$ and the mutant sentence $s'$.
In this case, a token or phrase is changed in the mutant sentence $s'$, which may result in a new named entity that does not exist in the original sentence.
Thus, we would not check the named entities that do not exist in both sentences, since the NER predictions of these named entities are not defined in both sentences.

% In this case, a token or phrase is changed in the mutant sentence $s'$ we would not check the named entities that do not exist in both sentences, since the NER predictions of these named entities are not defined in both sentences.
% \hyy{I think we need to explain the notation in more detail (The $e$ should be new words I think).}

However, structural transformation and random entity shuffle do not involve new named entities, which only change the order of the named entities or the syntactic structure of the sentence. 
Therefore, we check whether the test inputs satisfy the second metamorphic relation $MR_2$, defined as below:
\begin{equation}
    N(s) = N(s').
\end{equation}
If a test input does not satisfy the metamorphic relations, {\methodname} will report the test input as a suspicious issue.

\subsection{Automatically Repairing the Named Entity Recognition Errors }
After a suspicious issue is raised, we will repair both sentences $s$ and $s'$ to obtain fixed NER predictions $R(s)$ and $R(s')$.
For the original sentence $s$, if $N(s) \neq R(s)$, we will report a repairing attempt replacing $N(s)$ with $R(s)$, and for the mutant sentence $s'$, if $N(s') \neq R(s')$, we will report a repairing attempt replacing $N(s')$ with $R(s')$.
The process of black-boxed repairing consists of the following two parts, i.e., suspicious entity location and relabeling.

\begin{figure}
 \centering
 \includegraphics[width=1\linewidth]{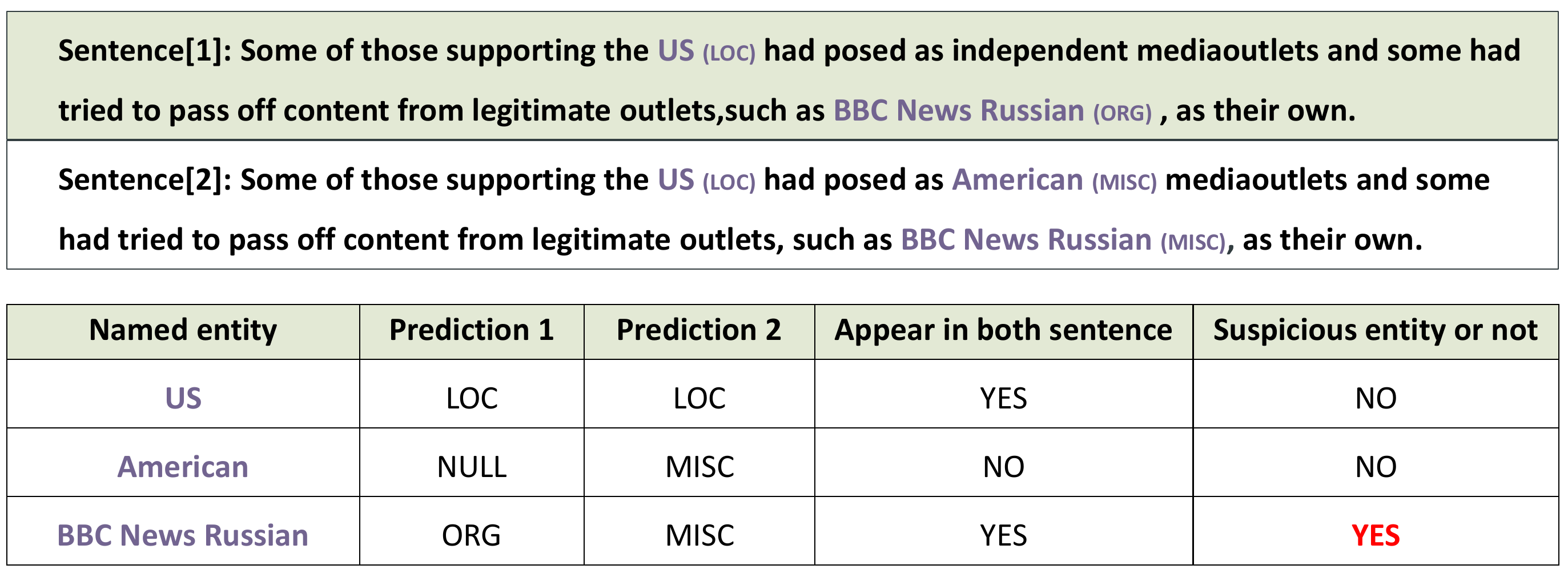}
 \caption{Suspicious entities location}
 \label{fig:location}
\end{figure}

\subsubsection{Suspicious Entity Location} 
A suspicious issue consists of two sentences and their NER predictions.
However, we do not know which named entity contains the NER errors and needs to be repaired.
To ensure the quality of the fixed NER prediction, we need to locate the suspicious entities that are prone to NER errors.
% i.e, the named entities which are prone to contain NER errors.
We compare the contents and the NER predictions between the original sentence $s$ and the mutant sentence $s'$ to locate suspicious entities.
For each sentence, an entity $e$ is a suspicious entity if both of the following conditions are satisfied.

\begin{itemize}
    \item The entity exists in both sentences: $e \in s \text{ and } e \in s'$
    \item The NER prediction of this entity is different between the two sentences: $N(s, e) \neq N(s', e)$
\end{itemize}
We then add all satisfied $e$ to a list as the suspicious entities of both sentences.

For example, as shown in Fig.~\ref{fig:location}, we first iterate all the named entities predicted by the NER systems and obtain their NER categories.
The information on these named entities is shown in the table.
"\textit{BBC News Russian}" appears in both sentences, and the predictions of its NER category are different between the two sentences, thus it is reported as a suspicious entity located.

\subsubsection{Suspicious Entity Relabeling}
Given a suspicious entity, we would relabel the suspicious entity by our repairing algorithm in turn (one suspicious entity being relabeled each time).
Similar to the automated testing part, we first generate and filter the mutant entities which have similar semantics to the suspicious entity.
Then we feed the mutant entities into a scoring system to obtain the fixed NER prediction.

\textbf{1) Similar entity generation: }Given an input sentence $s$ and a suspicious entity $e_s$, {\methodname} masks each word/subword (a fragment of a word that might not stand alone as a full word in the language) tokenized by $\text{BERT}$ of $e_s$ in turn (one word or subword being masked each time) and feeds the masked sentences into $\text{BERT\_MLM}$ to generate top-K mutant entities with predictive logit.

\textbf{2) Similar entity filter: } Next, to ensure high quality of the similar entities, we adopt another threshold $PThreshold$ and filter out mutant entities with a predictive logit $p$ lower than $PThreshold$.
To ensure the mutant entities have consistent semantics with the suspicious entity, we then compute the context-aware embedding for the suspicious entity and the mutant entities.
For each mutant entity $e_i$, we filter out $e_i$ if the predefined semantic similarity between the embedding of $e_s$ and embedding of $e_i$ is lower than a threshold $SThreshold$.
We also check the format consistency between the suspicious entity and the mutant entities, and remove those with inconsistent formats.
For example, the case of the first letter of the generated word must remain unchanged.
In practice, we set $PThreshold$ as 5.5 and $SThreshold$ as 0.45 based on experience, which outperforms well on the four NER systems.

\begin{figure}
 \centering
\includegraphics[width=1\linewidth]{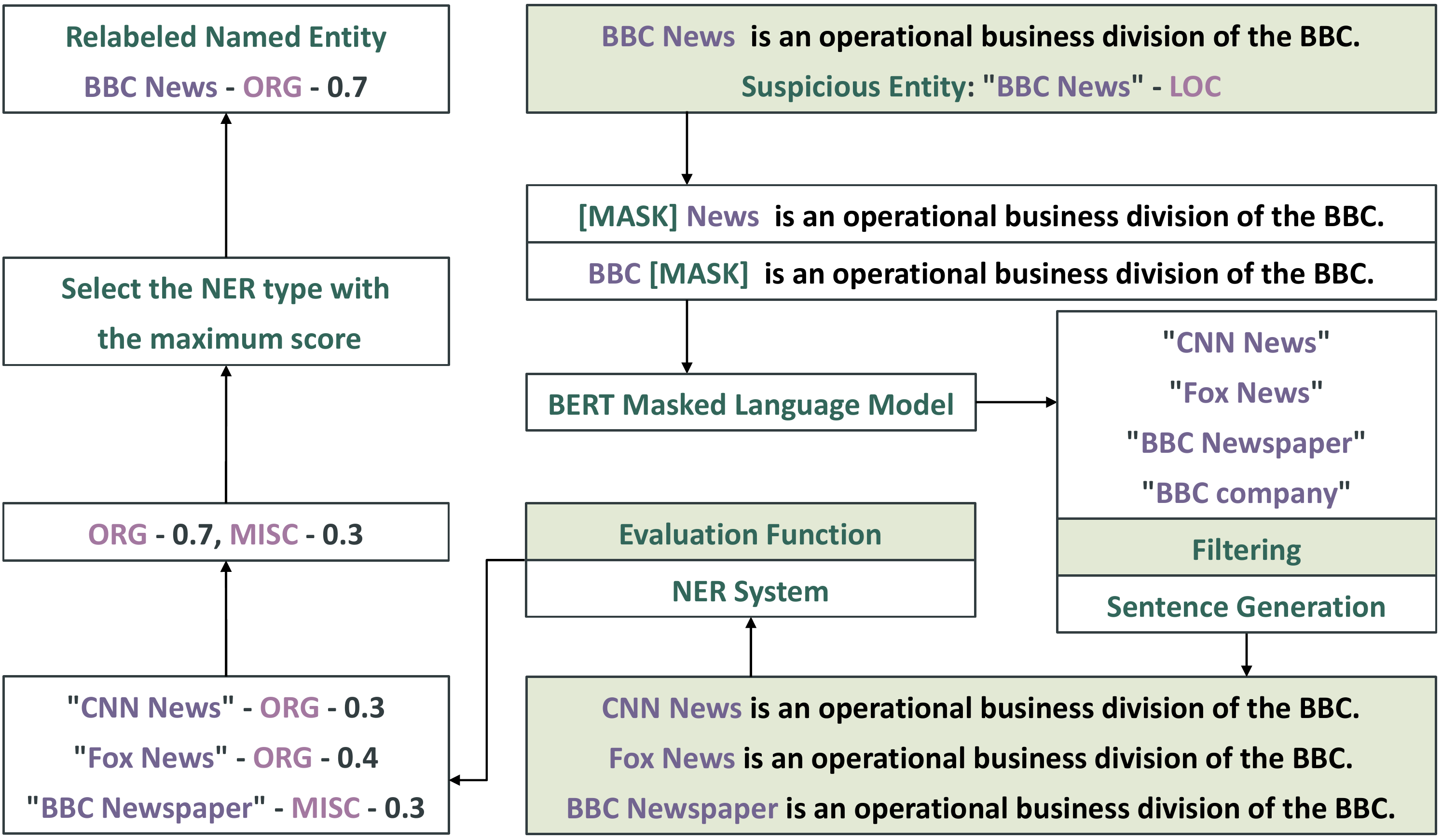}
 \caption{Suspicious entity relabel}
 \label{fig:scoreSystem}
\end{figure}

\begin{figure}
 \centering
\includegraphics[width=0.99\linewidth]{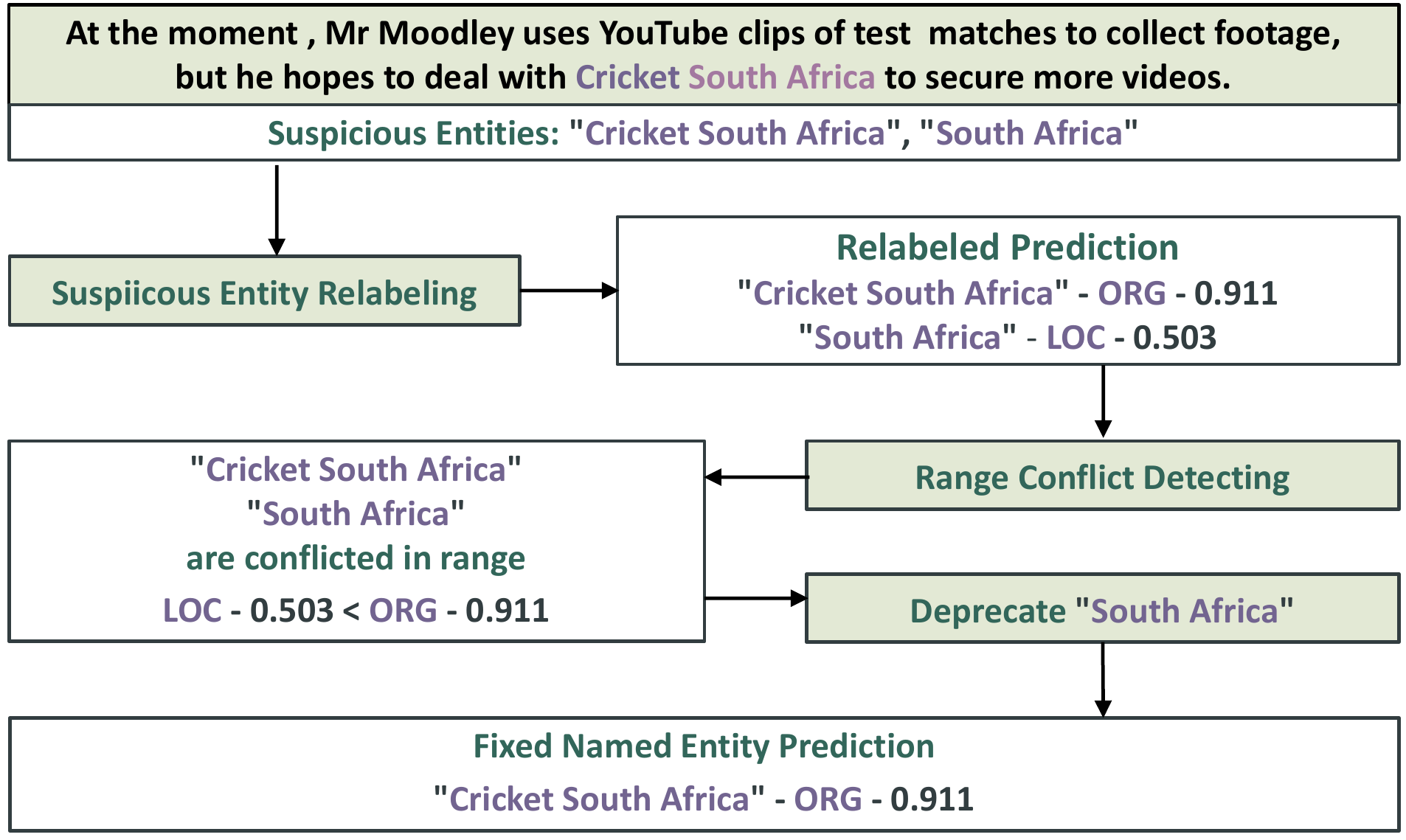}
 \caption{Repairing of range conflict}
 \label{fig:rangeConflict}
\end{figure}

\textbf{3) Scoring system: }After entity generation and filtering, we replace the suspicious entity with the filtered mutant entities $e'_1, e'_2, \dots\\ e'_{m}$ in turn to generate mutant sentences $s'_1,s'_2,\dots s'_m$, which will be fed into the NER system to obtain the NER prediction of each mutant entity $\forall_{1 \leq i \leq m} N(s'_i, e'_i) $.
Specifically, we add a new NER category named \texttt{NULL} to indicate that the words should not be recognized as a named entity.
For each NER category, we add a score calculated by an evaluation function $F(p, sim)$ to it in the scoring system, where $p$ denotes the predictive logit of the mutant entity and $sim$ denotes the semantic similarity.
For the masked word/subword $w$, mutant $e'$ and mutant sentence $s'$, we define the evaluation function as below:
\begin{small}\begin{equation}
F(p, sim) = 
    \begin{cases}
    \lambda \cdot \alpha \cdot p \exp(k \cdot sim) & N(s', e') = \texttt{NULL}, w \text{ is subword},\\
    \alpha \cdot p \exp(k \cdot sim) & N(s', e') = \texttt{NULL}, w \text{ is word},\\
    \lambda \cdot p \exp(k\cdot sim) & N(s', e') \neq \texttt{NULL}, w \text{ is subword},\\
    p \exp(k\cdot sim) & N(s', e') \neq \texttt{NULL}, w \text{ is word}
    \end{cases}
\end{equation}\end{small}

where $k$ is a constant to balance the weights between the predictive logit and the semantic similarity, $\alpha, \lambda$ are constant coefficients between $(0, 1]$ used for controlling the influence of the newly added NER category \texttt{NULL} and the subword tokenized by \text{BERT}, respectively.
% for the NER type of "NULL" to reduce the influence from the unnatural mutant sentences which is unfiltered.
In practice, we set $k = 2.5$, $\alpha = 0.2$, and $\lambda = 0.5$ for all NER systems we adopt {\methodname} to repair.
Finally, the NER category with the maximum score in the scoring system will be selected as the relabeled NER prediction of the suspicious entity.

The overall process of the relabeling is detailed by Algorithm~\ref{alg:relabel}. 
We define a function Relabel where the input consists of a suspicious entity $e_s$ and the sentence $s$.
The output of the function is the relabeled NER category $prediction$ and a score $p\_score$ for the NER category.
We first define an empty dict $score$ to store the score of each NER category during the process (Line 2).
Then we enumerate each word/subword $w_a$ in $e_s$ (Line 4) and mask $w_a$ by a special mark "[MASK]" to obtain a masked sentence $s_{mask}$ (Line 5).
The masked language model function $\mathrm{BERT\_MLM}$ generates a list of candidate words $C_w$ with the corresponding predictive logits $p$ and returns the top-K candidate words with the highest predictive logit (Line 6).
The embedding function $\mathrm{BERT\_Embedding}$ extracts the context-aware embedding of the suspicious entity and the mutant entities (Lines 3 and 12).
We enumerate each word $w_b$ and its predictive logit $p$ in $C_w$ (Line 7) and filter them through the introduced filters: format Consistency, predictive logit, and 
semantic similarity (Lines 8 to 14).
After that, we generate the mutant entity $e_m$ and the mutant sentence $s_m$ with the filtered $w_b$, which will be fed into the NER system to obtain the NER prediction of $e_m$ (including the newly added category \texttt{NULL}), denoted as $N(s_m, e_m)$.
Then we use the evaluation function $F(p, sim)$ to compute the score for the $e_m$, and add the score to $score[N(s_m,e_m)]$ (Line 15).
Finally, we select the NER category with the maximum score in the dict as the relabeled NER category $prediction$ (Line 18) and its score as $p\_score$ (Line 19).

There is an example using {\methodname} to relabel the suspicious entity "\textit{BBC News}" - \texttt{LOC} in the sentence "\textit{BBC News is an operational business division of the BBC.}", as shown in Fig.~\ref{fig:scoreSystem}.
{\methodname} first masks each word/subword in the "\textit{BBC News}" in turn to generate mutant entities "\textit{CNN News}", "\textit{Fox News}", "\textit{BBC Newspaper}", and "\textit{BBC company}" through the BERT. 
Then {\methodname} filters out the mutant entity "\textit{BBC company}" and feeds the mutant sentences into the NER system to obtain the NER predictions "\textit{CNN News}" - \texttt{ORG}, "\textit{Fox News}" - \texttt{ORG}, and "\textit{BBC Newspaper}" - \texttt{MISC}.
The evaluation function of {\methodname} will calculate the score for each filtered mutant entity and aggregate the score to the corresponding NER category: \texttt{ORG} ($score = 0.7$), \texttt{MISC} ($score = 0.3$).
After that, {\methodname} selects the NER category \texttt{ORG} that has the maximum score as the relabeled result of the "\textit{BBC News}".

\textbf{4) Repairing of range conflict: }As the relabeling process is independent for each suspicious entity, there are potential range conflicts among the relabeled named entities, which will be solved by {\methodname} using $p\_score$.
In practice, we enumerate all pair $(e_a, e_b)$ of relabeled named entity within $s$ and check whether there is a range conflict between $e_a$ and $e_b$.
If there is a range conflict among two relabeled named entities, {\methodname} will deprecate the relabeled named entity with lower $p\_score$ between $e_a$ and $e_b$.

For example, as shown in Fig~\ref{fig:rangeConflict}, there is a range conflict between two suspicious entities "\textit{Cricket South Africa}" and "\textit{South Africa}" in the sentence.
After the relabeling process, the $p\_score$ of the former is $0.911$ whereas the latter is $0.503$, by which {\methodname} will deprecate the relabeled named entity "\textit{South Africa}".

Finally, we obtain a fixed NER prediction $R(s)$ without range conflicts.
If the fixed NER prediction differs from the original $N(s)$, the NER system is repaired using $R(s)$.

% However, the relabeled results may have range conflicts. 
% For example, as shown in Fig. \mqy{Here need a pic and description for range conflict}
\begin{algorithm}\small
\caption{Black-box relabeling the suspicious entity\label{alg:relabel}} 
\KwData{$s$: a sentence input; $e_s$: the suspicious entity to relabel;}
\KwResult{$prediction$: the relabeled NER type of the suspicious entity; $p\_score$: the score of the relabeled NER type. }
\SetKwFunction{Relabel}{Relabel}
\SetKwProg{fn}{Function}{:}{}
\fn{\Relabel{$s$, $e_s$}}{
$score = \mathrm{Dict}()$ \\
$h_s = \mathrm{BERT\_Embedding}(s, e_s)$
\\
\For{each $w_a$ $\in$ $e_s$ }
{
    $s_{mask} = \mathrm{Replace}(s, w_a, \text{[MASK]})$ 
    \\
    $C_w = \mathrm{BERT\_MLM}(s_{mask})$
    \\
    \For{each $(w_b, p) \in C_w$}
    {
        \lIf{$\mathrm{Format\_Incosistent}(w_a, w_b)$}{
            \textbf{continue}
        }
        \lIf{$p < PThreshold$} 
        {
            \textbf{continue}
        }
        $s_m = \mathrm{Replace}(s, w_a, w_b)$
        \\
        $e_m = \mathrm{Replace}(e_s, w_a, w_b)$
        \\
        $h_m = \mathrm{BERT\_Embedding}(s_m, e_m)$
        \\
        $sim = \mathrm{CosSim}(h_s, h_m)$
        \\
        \lIf{ $sim < SThreshold$}
        {
            \textbf{continue}
        }
        $score[N(s_m, e_m)] \mathrel{+}= F(p, sim)$
    }

}
$prediction = \mathop{\arg\max}_{key \in \mathrm{NER\_Category}}(score[key])$
\\
$p\_score = \max(score.values)$
\\
\Return $prediction, p\_score$
}
\textbf{End Function}

\end{algorithm}

\begin{table*}[h]
    \centering
    \caption{Precision of {\methodname} on BBC News dataset}
    \resizebox{0.99\linewidth}{!}{
    \begin{tabular}{c|ccccc}
    \hline 
    NER systems & Token-level Replacement & Phrase-level Replacement & Structural Transformation  & Random Entity Shuffle & {\methodname} Overall   \\
    \hline 
    Flair-CoNLL & 86.0\% (43/50) & 90.0\% (45/50) & 91.7\% (33/36) & 80.0\% (40/50) & 86.6\% (161/186) \\
    Flair-Ontonotes & 82.0\% (41/50) & 90.0\% (45/50) & 90.0\% (45/50) & 78.0\% (39/50) & 85.0\% (170/200) \\
    Azure NER & 90.0\% (45/50) & 92.0\% (46/50) & 98.0\% (49/50) & 92.0\% (46/50) & 93.0\% (186/200) \\
    AWS NER & 96.0\% (48/50) & 96.0\% (48/50) & 93.8\% (45/48) & 88.0\% (44/50) & 93.4\% (185/198) \\
    
    % MS Azure API & 88.13  & 97.68 & 97.56 \\
    \hline
    \end{tabular}
    \label{tab:precision}
    }
\end{table*}

\begin{table*}[h]
    \centering
    \caption{Precision comparison between {\methodname} and baseline on CoNLL03 (tested on Flair-CoNLL)}
    \resizebox{1.00\linewidth}{!}{
    \begin{tabular}{c|ccccc|cc}
    \hline
     & & & {\methodname} & & & \ \ \ \ \ \ \ \ \ \ \ \ \ \ \ \ \ \ \ \ \ \ \ \ \ \ SeqAttack  \\
    \hline 
    Methods & Token-level Replacement & Phrase-level Replacement & Structural Transformation  & Random Entity Shuffle & {\methodname} Overall & BERT-Attack & Clare \\
    % \hline 
    Precision & 96.0\% (48/50) & 96.0\% (48/50) & 92.9\% (26/28) & 98.0\% (49/50) & 96.1\% (171/178) & 24.0\% (12/50) & 28.0\% (14/50) \\
  
    % MS Azure API & 88.13  & 97.68 & 97.56 \\
    \hline
    \end{tabular}
    \label{tab:cmpbaseline}
    }
\end{table*}
\section{Evaluation}

\begin{itemize}
\item{RQ1: How accurate is {\methodname} at finding erroneous issues?}
% \item{RQ2: How many named entity recognition errors can our approach report?}
\item{RQ2: What categories of named entity recognition errors can {\methodname} find?}
\item{RQ3: How effective is {\methodname} at repairing NER errors?}
% \item{RQ4: How many types of NER errors can
% ATINER repair? 
% }
\end{itemize}

\subsection{Experiment Setup}
% \subsection{Experiment Environment}
To verify the results of {\methodname}, we manually inspect the results both for the automated testing and the automated repairing part of {\methodname}.
For the automated testing part of {\methodname}, we collectively decide: (1) whether the issue contains NER errors; and (2) if yes, what category of NER errors it contains.
For the automated repairing part of {\methodname}, we collectively decide whether the suspicious entity has been repaired correctly.
It is noticeable that manual inspection is only used to verify the effectiveness of {\methodname}, which is totally automated.
Specifically, we evaluated 784 suspicious issues for NER testing and 3,828 suspicious entities for NER repairing.
All experiments are conducted on a Linux (Ubuntu 20.04.2 LTS) workstation with 64GB Memory and GeForce RTX 3090 GPU.
% \bx{Add the information of the package we use for processing the sentences.}\hyy{Do we still need this part? I pre-process the sentence with some python codes and in the testing part I use the stanford parser and nltk package to ensure the format is consistent}

% \usepackage{booktabs}

\subsection{Dataset}
We use two datasets in this paper, the BBC News dataset and CoNLL03 dataset.
We collect the sentences on BBC news with multiple news categories to construct the BBC News dataset, which contains 3,223 sentences without NER labels.
We use this dataset to evaluate {\methodname}'s performance to test and repair NER systems on general text data.
% Since {\methodname} does not need labels for testing and repairing NER systems, we collect the BBC News dataset without the NER labels to verify the effectiveness of {\methodname}.
The CoNLL03 dataset is a widely used NER dataset that contains the ground truth label.
As the baselines rely on ground truth label to generate test cases, we compare {\methodname} and the baselines on the test dataset of CoNLL03.

% \hyy{Maybe we need to mention that our evaluation is still independent of gt label of conll03 dataset. We may mention that the dataset has some problems in the discussion part}.
% We need to compare {\methodname} with the baselines that need the information of the named entities in the ground truth label to generate mutant sentences, thus we compare {\methodname} with the baselines on the test dataset of CoNLL03~\cite{sang2003introduction}.

\subsection{Comparison}
{\methodname} is the first black-box testing and repairing approach for general NER systems, and there is a dearth of baselines for testing approaches.
Alternatively, we choose two popular adversarial attack approaches as our baseline methods Bert-Attack~\cite{li2020bert} and Clare~\cite{li2020contextualized}, and compare {\methodname} with the baseline methods on the CoNLL03 test dataset~\cite{sang2003introduction}.
% \hyy{It is strange to use "therefore", as there is no reason and consequence logic in the preceding sentences. I think we can say that the adversarial attack models are somehow similar to our model (both try to find bugs). By comparing the baseline with our model, we can check how effective our model is in terms of finding errors}
We use {\methodname} and the baselines to test the Bert-Base-NER~\cite{devlin2018bert} fine-tuned on CoNLL03 dataset, and evaluate the precision of the reported suspicious issues.

\subsection{Evaluation Metric}\label{sec:eval}
As {\methodname} consists of automated testing and automated repairing for NER systems, we use different evaluation metrics for these two parts.
For NER automated testing, the output is a list of suspicious issues, each containing (1) original sentence $s$ and its NER predictions $N(s)$, (2) mutant sentence $s'$ and its NER predictions $N(s')$.
We define the Precision of NER automated testing as the percentage of erroneous issues in suspicious issues, which have NER prediction errors in at least one sentence.
We use Precision to evaluate how precise {\methodname} reports the suspicious issues.
Explicitly, for a suspicious issue $p_t$, we set $error_t(p_t)$ to $\mathrm{True}$ if $N(s)$ or $N(s')$ have NER prediction errors.
Otherwise, we will set $error_t(p_t)$ to $\mathrm{False}$.
Given a list of suspicious issues, the $\mathrm{Precision}$ is calculated by:
\begin{equation}
\label{equ:precision}
\mathrm{Precision} = \frac{\sum_{p_t \in P_T}  \mathds{1}\{error(p_t)\}}{|P_T|},
\end{equation}
where $P_T$ is the suspicious issues returned by {\methodname} and $|P_T|$ is the number of suspicious issues from automated NER testing.
% \hyy{I think we may define what is an "error", eg: we may define a ground truth model $N_{gt}$ such that $N(s) \neq N_{gt}(s)$}

For automated NER repairing, there are four possible outcomes: $TF, FT, FF, TT$.
We use $T$ and $F$ to represent whether the NER prediction is correct or incorrect, and $TF, FT, FF, TT$ present the transition through the repairing, e.g., $FT$ means the repairing process successfully fixes the NER systems by changing incorrect prediction to correct.
The number of incorrect NER predictions before repairing is $NumError=FT+FF$ and the number of correct NER predictions before repairing is $NumCorrect=TT+TF$. Three evaluation metrics are used to evaluate the effectiveness of NER repairing:

% The evaluation metrics are defined below:
\begin{itemize}
\item$Err2Cor$ measures the probability of changing incorrect NER predictions to correct and is calculated as $Err2Cor = \frac{FT}{NumError}$.

\item $Cor2Err$ measures the probability of changing correct NER predictions to incorrect and is calculated as $Cor2Err =\\ \frac{TF}{NumCorrect}$.

\item $ErrorReduce$ measures the ability to reduce NER errors and is calculated as $ErrorReduce = \frac{FT-TF}{NumError}$.
%\end{itemize}
% \begin{itemize}
% \item $Err2Cor$: the ratio measuring the number of NER errors being repaired among the number of incorrect NER predictions before repairing, which is calculated by: $Err2Cor = \frac{FT}{NumError}$.
% \item $Cor2Err$: the ratio between the number of suspicious entities that {\methodname} changing correct NER prediction to incorrect and the number of suspicious entities with correct NER prediction, \textcolor{blue}{which measures the probability that {\methodname} fails and turns the correct NER predictions into wrong:} $Cor2Err = \frac{TF}{NumCorrect}$.
% \item $ErrorReduce$: the ratio between the number of NER errors reduced after the repairing procedure and the number of total NER errors detected, \textcolor{blue}{which measures the ability that {\methodname} reduces the NER errors:} $ErrorReduce = \frac{FT-TF}{NumError}$.
\end{itemize}
\subsection{RQ1: Precision of Finding Erroneous Issues}
We adopt {\methodname} to test four NER systems and report suspicious issues, where the effectiveness lies in how precise the reported issues are.
The evaluation result on BBC News Dataset is shown in Table~\ref{tab:precision}, where {\methodname} achieves a high average precision ranging from 85.0\% to 93.4\% over the four NER systems under test.
Among all the transformation methods on the four NER systems under test, {\methodname} achieves a high precision from 78.0\% to 98.0\%.
On BBC News Dataset, {\methodname} successfully reports 702 erroneous issues out of 784 suspicious issues we sample.
% \hyy{should I highlight the highest precision in each row of the table?}
The evaluation result on BBC News Dataset has shown that {\methodname} achieve a good performance on the dataset without ground truth label.
We also compare {\methodname} with the baselines on ConLL03, where the result is shown in Table~\ref{tab:cmpbaseline}
% \bx{wait for yanger's figure}.
We can observe that {\methodname} achieves much higher precision than the baselines, ranging from 92.9\% to 98.0\%.
While BERT-Attack and Clare only achieve a low precision of 24.0\% and 28.0\%, respectively.
Therefore, {\methodname} has a significant advantage over the baselines for reporting NER errors, even though it does not need the ground truth label to generate the test cases.
% \hyy{The "even though" part here is kind of strange. I think we do not need the part after "even though". We just need to highlight how precise our model can be. In addition, should we mention how we label the suspicious cases? eg: train and hire people to check the issues. I think we may create a subsection to discuss about the human-labeling job}
% \begin{figure*}
%  \centering
%  \includegraphics[width=0.99\linewidth]{Images/baseline.png}
%  \caption{Precision comparison between ANTER and baseline on CoNLL03}
%  \label{fig:baseline}
% \end{figure*}

\begin{center}
\fbox{
\parbox{0.963\linewidth}{\textbf{RQ1 Answer}: {\methodname} achieves a high average precision on the BBC News Dataset without ground truth label, ranging from  85.0\% to 93.4\% over the four NER systems under test.
For the comparison between {\methodname} and the baselines on CoNLL03, {\methodname} has achieved a much higher precision than the baselines.
}
}
\end{center}

% \subsection{Erroneous named entity recognition}

\begin{table*}[h]
\caption{Example of four categories of NER errors and repairing of them by {\methodname}}
\centering
\resizebox{1.0\linewidth}{!}{
\begin{tabular}{@{}lccc@{}}
\toprule
\multicolumn{1}{c}{\textbf{Sentence}} &
  \textbf{Suspicious Entities} &
  \textbf{Original Prediction} &
  \textbf{Fixed Prediction} \\ \midrule
\begin{tabular}[c]{@{}l@{}}Ben Johnson , from the Environmental Services Association (\textcolor{red}{ESA}), \\ told BBC News more and more people were putting devices \\ containing these batteries in with household rubbish or \\ mixing them with other recycling. (\textbf{Flair-CoNLL})\end{tabular} &
  {[}"ESA"{]} &
  \begin{tabular}[c]{@{}c@{}}{[}"Ben Johnson", \texttt{PER}{]} \\ {[}"Environmental Services Association", \texttt{ORG}{]}\\ {[}"BBC News", \texttt{ORG}{]}\\ \textcolor{red}{Error Category: \textit{Omission}}\end{tabular} &
  \begin{tabular}[c]{@{}c@{}}{[}"Ben Johnson", \texttt{PER}{]}\\ {[}"Environmental Services Association", \texttt{ORG}{]}\\ {[}"BBC News", \texttt{ORG}{]}\\ {[}\textcolor{blue}{"ESA", \texttt{ORG}}{]}\end{tabular} \\ \midrule
\begin{tabular}[c]{@{}l@{}}The \textcolor{red}{halfway} point affords us an opportunity to step back and \\ then look at what our margins are and where we could be a \\ little smarter to buy down risk and understand the spacecraft's \\ performance for crewed flight on the very next mission. (\textbf{Flair-Ontonotes})\end{tabular} &
  {[}"halfway"{]} &
  \begin{tabular}[c]{@{}c@{}}{[}\textcolor{blue}{"halfway", \texttt{CARDINAL}}{]}\\ \textcolor{red}{Error Category: \textit{Over-labeling}}\end{tabular} & \begin{tabular}[c]{@{}c@{}}{[}\textcolor{blue}{"halfway", \texttt{NULL}}{]}\end{tabular} \\ \midrule
\begin{tabular}[c]{@{}l@{}}They say the only positive thing the federal authorities have done \\ is to return electricity to \textcolor{red}{Mekelle}. (\textbf{Azure NER})\end{tabular} &
  {[}"Mekelle"{]} &
  \begin{tabular}[c]{@{}c@{}}{[}"authorities", \texttt{PERSONTYPE}{]}\\ {[}\textcolor{blue}{"Mekelle", \texttt{PERSON}}{]}\\ \textcolor{red}{Error Category: \textit{Incorrect Category}}\end{tabular} &
  \begin{tabular}[c]{@{}c@{}}{[}"authorities", \texttt{PERSONTYPE}{]}\\ {[}\textcolor{blue}{"Mekelle", \texttt{LOCATION}}{]}\end{tabular} \\ \midrule
\begin{tabular}[c]{@{}l@{}}Fibrus is delivering a similar scheme in Northern Ireland known as \\ \textcolor{red}{Project Stratum}. (\textbf{AWS NER})\end{tabular} &
  \begin{tabular}[c]{@{}c@{}}{[}"Project"{]}\\ {[}"Stratum"{]}\\ {[}"Project Stratum"{]}\end{tabular} &
  \begin{tabular}[c]{@{}c@{}}{[}"Fibrus", \texttt{ORGANIZATION}{]}\\ {[}"Northern Ireland", \texttt{LOCATION}{]}\\ {[}\textcolor{blue}{"Project", \texttt{OTHER}}{]}\\ {[}\textcolor{blue}{"Stratum", \texttt{TITLE}}{]}\\ \textcolor{red}{Error Category: \textit{Range Error}}\end{tabular} &
  \begin{tabular}[c]{@{}c@{}}{[}"Fibrus", \texttt{ORGANIZATION}{]}\\ {[}"Northern Ireland", \texttt{LOCATION}{]}\\ {[}\textcolor{blue}{"Project Stratum", \texttt{OTHER}}{]}\end{tabular} \\ \bottomrule
\end{tabular}
\label{tab:repair_type}
}
\end{table*}

\begin{table*}[h]
\caption{ Improvement based on manual inspection  }
\centering
\resizebox{1\linewidth}{!}{
\begin{tabular}{c|cccccc|ccc}
\hline
\textbf{NER Systems} &
  \textbf{$TT$} &
  \textbf{$TF$} &
  \textbf{$FT$} &
  \textbf{$FF$} &
  \textbf{$NumError$} &
  \textbf{$NumCorrect$} &
  \textbf{$Err2Cor$} &
  \textbf{$Cor2Err$} &
  \textbf{$ErrorReduce$} \\ \hline
Flair-ConNLL & 286 ({41.4\%}) & 48 ({7.0\%}) & 192 ({27.8\%}) & 164 ({23.8\%}) & 356 (51.6\%) & 334 (48.4\%) & 53.9\% & 14.4\% & \textbf{40.4\%} \\
Flair-Ontonotes &
  \multicolumn{1}{l}{483 ({42.7\%})} &
  \multicolumn{1}{l}{117 ({10.3\%})} &
  \multicolumn{1}{l}{264 ({23.3\%})} &
  \multicolumn{1}{l}{285 ({25.2\%})} &
  549 (47.8\%) & 600 (52.2\%) &
  48.1\% & 19.5\% & \textbf{26.8\%} \\
AWS NER & 456 ({44.8\%}) & 63 ({6.2\%}) & 275 ({27.0\%}) & 223 ({21.9\%}) & 498 (49.0\%) & 519 (51.0\%) & 55.2\% & 12.1\% & \textbf{42.6\%} \\
Azure NER & 413 ({42.5\%}) & 85 ({8.7\%}) & 325 ({33.4\%}) &  149 ({15.3\%})  & 474 (48.8\%) & 498 (51.2\%)
 & 68.6\% & 17.1\% & \textbf{50.6\%} \\ \hline
\end{tabular}
\label{tab:repair_performance}
}
\end{table*}

\subsection{RQ2: Categories of the Reported NER Errors}
{\methodname} is capable of finding NER errors of diverse kinds.
In our experimental results, we mainly found 4 categories of NER errors: omission, over-labeling, incorrect category, and range error, where the categories are concluded during our manual inspection.
% \bx{We need a figure to show the distribution of each type of NER error.}
We will show examples of the reported NER errors with respective to each category of NER error in the following.

\noindent \textbf{Omission}
An omission error indicates that the NER systems fail to recognize the named entity in a sentence.
For example, in the first sentence of Table~\ref{tab:repair_type}, Flair-CoNLL fails to detect the named entity "\textit{ESA}".

\noindent \textbf{Over-labeling}
An over-labeling error indicates that the NER systems label the words that do not belong to any NER category.
For example, in the second sentence of Table~\ref{tab:repair_type}, Flair-Ontonotes mistakenly label the "\textit{halfway}" as the NER category \texttt{CARDINAL}.

\noindent \textbf{Incorrect category}
An incorrect category error indicates that the NER systems incorrectly predict the NER category of the named entity.
For example, in the third sentence of Table~\ref{tab:repair_type}, Azure NER predicts "\textit{Mekelle}" as \texttt{PERSON}, which is indeed a \texttt{LOCATION}.

\noindent \textbf{Range error}
A range error indicates the NER systems only label part of the named entity or involve unnecessary elements besides the named entity.
For example, in the fourth sentence of Table~\ref{tab:repair_type}, AWS NER erroneously predicts the two words "\textit{Project}" and "\textit{Atratum}" of the named entity "\textit{Project Atratum}" (a specific project name which belongs to \texttt{OTHER} in AWS NER categories) as \texttt{OTHER} and \texttt{TITLE}, which overlooks the whole meaning of the named entity and leads to the range error.

We calculate the distribution of the four NER categories on a random sample of 468 NER errors found by {\methodname}, where omission, over-labeling, incorrect category, and range error, each accounts for 16.9\% (79/468), 19.6\% (92/468), 34.2\% (160/468), 29.3\% (137/468).

% \bx{Yan add the examples in the form of the figure here}
% \hyy{1. Should we add a figure to show the error distribution? I think we may also show the situation of the problematic sentences (some sentences may have more than one error). It is interesting to see the distribution of errors over erroneous sentences. We can use a Venn graph to show such distribution. 3. I think we can analyze the distribution of errors on different APIs}

% \vspace{1}
\begin{center}
\fbox{
\parbox{0.963\linewidth}{\textbf{RQ2 Answer}: {\methodname} can successfully find four categories of NER errors, including omission, over-labeling, incorrect category, and range error, which occupies 16.9\% 19.6\%, 34.2\% and 29.3\% on an evaluation of 468 NER errors.
}
}
\end{center}

\subsection{RQ3: Repairing Performance of {\methodname}}
We evaluate a total of 3,828 suspicious entities for assessing {\methodname}'s ability to repair the NER errors.
The results of the automated repairing are shown in Table~\ref{tab:repair_performance}, where we can observe that a big ratio of the suspicious entities contains NER errors, from 47.8\% to 51.6\% among the four NER systems under test.
We have found 356, 549, 498, and 474 NER errors for Flair-CoNLL, Flair-Ontonotes, AWS NER, and Azure NER, respectively.
By utilizing {\methodname} to repair these NER errors, 192/356, 264/549, 275/498, and 325/474 of the NER errors have been repaired.
The $Err2Cor$ in Table~\ref{tab:repair_performance} is used to show {\methodname}'s ability to repair the NER errors, where a high ratio from 48.1\% to 68.6\% of the NER errors have been repaired to be correct NER predictions among the four NER systems under test.
However, {\methodname} may sometimes turn the correct NER prediction into an incorrect prediction.
The $Cor2Err$ in Table~\ref{tab:repair_performance} is used to assess the extent where {\methodname} would mislead the NER systems.
We can observe that only 12.1\% to 19.5\% of the correct NER predictions have been misled to the incorrect NER predictions, which is much lower than the ratio that the NER errors are repaired.
$ErrorReduce$ considers both the errors being repaired and the correct NER predictions being misled.
We can observe that there is a great ratio of error reduction after the automated repairing, from 26.8\% to 50.6\%.
Therefore, we can see that the automated repairing effectively decreases the NER errors of the four NER systems under test.

As shown in Table~\ref{tab:repair_type}, {\methodname} can repair four categories of NER errors it finds:
% \vspace{-1.5mm} 

% \subsubsection*{\textbf{Repairing of Omission Error:}}
\noindent \textbf{Repairing of Omission Error:}
In the first sentence of Table~\ref{tab:repair_type}, {\methodname} locates the suspicious entity "\textit{ESA}" which is an abbreviation of the organization name "\textit{Environmental Services Association}".
In the original prediction, the NER system misses the NER prediction of "\textit{ESA}", where there is an error of omission.
{\methodname} successfully predicts the correct NER category \texttt{ORG} of "\textit{ESA}" in the fixed prediction.

% \vspace{-1.5mm} 

% \subsubsection*{\textbf{Repairing of Over-labeling Error:}}
\noindent \textbf{Repairing of Over-labeling Error:}
In the second sentence of Table~\ref{tab:repair_type}, {\methodname} locates the suspicious entity "\textit{halfway}", which does not belong to any NER category under the rule of Flair-Ontonotes.
In the original prediction, the NER system recognizes the NER prediction of "\textit{halfway}" as \texttt{CARDINAL},
where there is an error of over-labeling.
{\methodname} successfully deprecates the NER prediction of "\textit{halfway}" in the fixesd prediction.

% \vspace{-1.5mm} 

% \subsubsection*{\textbf{Repairing of Incorrect Category:}}
\noindent\textbf{Repairing of Incorrect Category:}
In the third sentence of Table~\ref{tab:repair_type}, {\methodname} locates the suspicious entity "\textit{Mekelle}", which is a special zone and capital of the Tigray Region of Ethiopia.
In the original prediction, the NER system recognizes the NER prediction of "\textit{Mekelle}" as incorrect NER category \texttt{PERSON},
where there is an error of incorrect category.
{\methodname} successfully predicts the correct NER category \texttt{LOCATION} of "\textit{Mekelle}" in the fixed prediction.

% \vspace{-1.5mm}

% \subsubsection*{\textbf{Repairing of Range Error:}}
\noindent\textbf{Repairing of Range Error:}
 In the fourth sentence of Table~\ref{tab:repair_type}, {\methodname} locates the suspicious entities "\textit{Project}", "\textit{Stratum}" and "\textit{Project Stratum}", where "\textit{Project Stratum}" is a broadband infrastructure project to extend access to superfast broadband services across Northern Ireland.
 In the original prediction, the two elements of "\textit{Project Stratum}", i.e., "\textit{Project}" and "\textit{Stratum}", are recognized as \texttt{OTHER} and \texttt{TITLE}, where there is an error of range.
{\methodname} successfully predicts the correct NER category \texttt{OTHER} of "\textit{Project Stratum}" and deprecates the NER predictions of "\textit{Project}" and \textit{Stratum} in the fixed prediction.

% To show that the NER systems have been improved through automated repairing, we also calculate $ErrorReduce$, which shows the ratio of NER errors being 

\begin{center}
\fbox{
\parbox{0.963\linewidth}{\textbf{RQ3 Answer}: 
The evaluation results show the great ability of {\methodname} to repair the NER systems, where $Err2Cor$ is much higher than $Cor2Err$, and there is a high ratio of $ErrorReduce$ from 26.8\% to 50.6\% on the four NER systems under test.
{\methodname} can repair four categories of NER errors it finds.
% The naturalness score of {\methodname} shows that its synthesized test cases are very close to natural images.}
}}
\end{center}
% \subsection{Effectiveness of naturalness filter (RQ4)}

% \subsection{ How many types of NER errors can {\methodname} repair? (RQ4) }

% {\methodname} is capable of repairing all four types of NER errors, thus improving the ability of the NER systems.
% We give four examples for repairing each type of NER error as shown in Table~\ref{tab:repair_type}, and explain how they repair the NER errors 
% in the four sentences as the following:

% \begin{itemize}
%     \item In the first sentence of Table~\ref{tab:repair_type}, {\methodname} locate the suspicious entity "ESA" which is an abbreviation of the organization name "Environmental Services Association".
%     In the original prediction, the NER system misses the NER prediction of "ESA".
%     {\methodname} successfully predicts the correct NER type "ORG" of "ESA" in the fixed prediction.
%     \item In the second sentence of Table~\ref{tab:repair_type}, 
%     \item In the third sentence of Table~\ref{tab:repair_type}, {\methodname} found that "Mekelle" is incorrectly recognized as "PERSON", where there is an error of incorrect category, then {\methodname} predict its correct tag "LOCATION".
%     \item In the fourth sentence of Table~\ref{tab:repair_type}, 
% \end{itemize}

% \begin{center}
    
% \fbox{
% \parbox{0.9\linewidth}{\textbf{RQ4 Answer}: 
% {\methodname} can successfully locate and repair four categories of Errors and give the correct prediction.
% }
% }

% \end{center}

\section{Discussion}

\subsection{False Positives}

% While {\methodname} tests the NER systems at high precision, there are still some false positives.
In the manual inspection of the experimental results for {\methodname}'s testing part, false positives refer to the suspicious issues that do not have NER errors.
In addition, if only the mutant sentence in a suspicious issue contains the NER errors while it is "unnatural", i.e., grammatically incorrect or difficult to understand, we also regard the suspicious issue as false positive. 
This is because NER errors detected on the "unnatural" mutant sentence contribute little to improving real-world NER systems.
In practice, we use a semantic filter and a syntactic filter to reduce the amounts of "unnatural" mutant sentences.

It is worth noting that the accuracy of NER systems does not impact the false positives of {\methodname}'s testing part, even though we use the NER prediction of the original sentence from NER systems to generate mutant sentences.
Specifically, if the original sentence's NER prediction is incorrect, it may lead to unexpected sentence transformation, e.g., replacing named entity in similar sentence generation.
However, when {\methodname} reports suspicious issues in this case, it would not result in false positives since the suspicious issues contain at least an original NER prediction error.

In the manual inspection of the experimental results for {\methodname}'s repairing part, false positives represent the case that {\methodname} turns the correct NER prediction to incorrect NER prediction, which is pre-defined as $TF$ in Section~\ref{sec:eval}.
However, $TF$ only ranges from 6.2\% to 10.3\% on the repairing results, which is much lower compared with the true positive indicator $FT$.
Despite the presence of false positives, {\methodname} can improve the performance of NER systems, reducing errors from 26.8\% to 50.6\%.
% With {\methodname}'s high precision to report errorneous issues and   
% For example, the NER system omits \textit{"Apple" - \texttt{ORG}} in the original sentence \textit{"Apple is a famous company"}.
% {\methodname} replace ""

% \hyy{1. We may use an example here. 2. This case is not an FP, it is strange to write this case in FP section.}

\subsection{Threat to Validity}
The main threats to external validity lie in the selection of the NER systems and the dataset for testing.
There are over hundreds of NER systems designed for multiple purposes and mountains of text data on the Internet, thus we can not evaluate {\methodname} on all of them.
To enrich the diversity of the test data, we collect the BBC News Dataset containing multiple news categories.
In addition, we evaluate {\methodname} on two typical SOTA NER systems in Flair~\cite{flairAPI}, and two famous commercial NER APIs to verify its generalization ability.
Threats to internal validity are factored into our experimental methodology, and they may affect our results.
Even though {\methodname} automatically tests and repairs the NER systems, we need to manually inspect the results to assess the performance of {\methodname}, which is potentially error-prone.
To minimize this threat, two people performed manual inspection separately and collectively decided whether {\methodname} succeeded in reporting erroneous issues and repairing NER errors.

% \subsubsection{Internel Validity}
% % While {\methodname} accurately tests the NER systems, its precision can be further removed.
% We have used multiple NLP tools for {\methodname}, including Stanford CoreNLP constituency parser~\cite{stanford_corenlp}, and NLTK's POS tagging tool~\cite{nltkAPI}, BERT~\cite{devlin2018bert}.
% However, these tools may give erroneous output under some circumstances, which may result in unnatural mutant sentences and false positives.
% The performance of {\methodname} could be further improved with the development of these NLP tools.

% \subsubsection{External Validity}
% The transformation approaches and the corresponding metamorphic relations are designed for English NER systems, which can not be directly used for the NER systems of other languages since the rules of grammar vary among different languages.

% \subsection{Building Robust NER Software}

\section{Related Work}

\subsection{Robustness of NLP Systems}
Deep neural networks have boosted the performance of multiple NLP tasks, including sentiment analysis~\cite{Iyyer18NAACL, Alzantot18EMNLP, Li19NDSS}, code analysis~\cite{Iyer16ACL, Pradel18OOPSLA, Alon19POPL}, reading comprehension~\cite{Chen16ACL, Chen18Thesis}, and machine translation~\cite{he2020structure, gupta2020machine, he2021testing, sun2020automatic, sun2022improving}.
However, there are still many bugs during the usage of the SOTA NLP systems.
% Inspired by the robustness studies in computer vision, 
Many researchers researched on the robustness of NLP systems, which unveiled bugs proposed by the neural networks used for various NLP systems~\cite{Jia17EMNLP, Caswell15TR, Miyato17ICLR, Iyyer18NAACL, Alzantot18EMNLP,Li19NDSS,Mudrakarta18ACL,Ribeiro18ACL}.
For example, Jia et al.~\cite{Jia17EMNLP} proposed an adversarial evaluation scheme for the Stanford Question Answering Dataset (SQuAD), which found seventeen SOTA models' performance is primarily undermined by the adversarial sentence inserted into the paragraph.
Similar to these works, {\methodname} focuses on the validation and improvement of the robutsness for NER systems.
Although some of {\methodname}'s mutation operators have already been utilized in existing NLP testing approaches~\cite{sun2020automatic, he2020structure} (e.g., word replacement), these established NLP testing tools cannot be seamlessly adapted for testing NER systems.
This is due to the need to avoid substituting named entities with non-named entities, and the absence of named entity labels in real-world text.
However, {\methodname} leverages the predictions of NER systems and performs mutations accordingly, eliminating the dependence on named entity labels for test case generation.
{\methodname} approach effectively addresses the challenges posed by traditional methods.
Furthermore, the testing framework employed by {\methodname} can be adopted to test other NLP systems, offering the advantage of finely controlling the components of the generated test cases.

% Although some of {\methodname}'s mutation operators have already been used in some existing NLP testing approaches~\cite{sun2020automatic, he2020structure} (e.g., word replacement), the existing NLP testing tools can not be easily adopted for testing NER systems.
% This is because we should avoid replacing the named entities with non-named entities, and there are not labels of named entities in the real-world text.
% However, {\methodname} utilizes the prediction of the NER systems and mutate based on it, which do not rely on the named entities labels to generate test cases and alleviate the challenges.

% He et al.~\cite{he2021testing} adopted referentially transparent input (RTI), which assumes the RTIs should have similar translations across different contexts.
% RTI has been used to report diverse kinds of machine translation errors.
% Different from these NLP systems, testing NER systems is quite challenging due to the diversity of the NER categories in different NER systems.
% Xu et al.~\cite{xu2022metamorphic} conducted a case study on Litigant, an industrial NER system by using metamorphic testing, but it can not be used to test other NER systems conveniently.
% Therefore, there is a dearth of automated testing approaches for general NER systems, and we designed {\methodname} to tackle this problem.

\subsection{Automated Improvement of NLP Systems}
Beyond the attack and test, improving the performance of the NLP systems is the ultimate goal.
Some researchers~\cite{gupta2020machine, he2020structure, he2021testing} relabeled the reported bugs of the NLP systems and fine-tuned the model on the relabeled dataset to fix the bugs.
Though effective, this method demands a significant amount of human resources.
Other researchers~\cite{sun2020automatic, sun2022improving} developed automated repairing techniques to fix the machine translation bugs.
For example, CAT~\cite{sun2022improving} and TransRepair~\cite{sun2020automatic} are word replacement methods that provide automatic fixing of revealed bugs without model retraining.
Inspired by CAT and TransRepair, we design a repairing system to fix the NER errors found by {\methodname}, which is the first black-box repairing algorithm to repair the general NER systems.

\subsection{Robustness of NER Systems}
Although NER systems have gained significant progress and benefited from deep neural networks, it is not a solved task yet.
Some researchers use adversarial methods~\cite{lin2021rockner, simoncini2021seqattack} to attack the NER systems, which undermines the performance of the NER systems under the perturbed datasets.
The adversarial attack method needs to access the parameters of the NER models, thus it can not be conveniently adopted to test the commercial NER APIs.
Xu et al.~\cite{xu2022metamorphic} adopted two specialized metamorphic relations to test an industrial NER system called Litigant.
% , which identifies \texttt{enterprise} and \texttt{person}, and their \texttt{role} from court judgment documents.
This approach can only be used to test Litigant because it relies on the characteristics (e.g., plaintiff and defendant roles in two companies) and application contexts (e.g., the transformation between parent and sub-company names via extension) of Litigant.
Thus, we did not include it in our experiments.
We proposed {\methodname}, the first black-box approach for testing and repairing general NER systems, which can be easily adapted to test various NER systems with different NER categories.

\subsection{Metamorphic Testing}
Metamorphic testing is a technique for testing software systems where the input and output of a test case are related in some way, such that if the input is modified in a specific way, the output should also be modified in a specific way.
As an effective approach to address the test case generation problems and test oracle problems, metamorphic testing is widely adopted to test a variety of systems, including: (1) traditional software, such as compilers~\cite{Le14PLDI, Lidbury15PLDI}, and database systems~\cite{lindvall2015metamorphic}, and (2) AI-based systems, such as machine translation~\cite{sun2020automatic, sun2022improving, he2020structure, gupta2020machine} and image captioning systems~\cite{yu2022automated}.
{\methodname} is the first metamorphic testing approach for validating general NER systems with different NER categories.

\section{Conclusion}
This paper presents {\methodname}, the first approach that automatically tests and improves the general NER systems, which is black-box and widely applicable to various NER systems.
We have proposed three transformation schemes to generate test cases and two metamorphic relations to test the NER systems, which achieve a high average precision from 85.0\% to 93.4\% for reporting erroneous issues over the four NER systems under test.
{\methodname} have successfully found a diversity of NER errors among the erroneous issues, including omission, mislabeling, incorrect category, and range error.
The automated repairing of {\methodname} can fix all four categories of NER errors and achieve a high rate of $ErrorReduce$ from 26.8\% to 50.6\% on the four NER systems under test.

% As a black-box method and does not rely on the ground truth label to generate test cases, {\methodname} is widely applicable to improve the robustness of various NER systems.

\section{Data Availability}
Codes and data of {\methodname} can be found at~\cite{tinCode}.

\begin{acks}
We thanks the anonymous ESEC/FSE reviewers and Xinwen Zhang for their valuable feedback on the earlier draft of this paper.
This paper was supported by the National Natural Science Foundation of China (No. 62102340) and Shenzhen Science and Technology Program.
\end{acks}

\balance

% \begin{acks}
% \end{acks}

\bibliographystyle{ACM-Reference-Format}
\bibliography{software}


\end{document}